\documentclass[review]{elsarticle}

\usepackage[hidelinks]{hyperref}
\usepackage{float}
\usepackage{algpseudocode}
\usepackage[tableposition=top]{caption}
\usepackage{amsmath}
\usepackage[ruled,linesnumbered]{algorithm2e}
\usepackage{graphicx}
\usepackage{subcaption}
\usepackage{caption}
\usepackage{setspace}

\makeatletter
\def\ps@pprintTitle{%
   \let\@oddhead\@empty
   \let\@evenhead\@empty
   \let\@oddfoot\@empty
   \let\@evenfoot\@oddfoot
}
\makeatother
\journal{Sustainable Cities and Society}

\begin{document}
% \linenumbers

\begin{frontmatter}

\title{Optimal Traffic Relief Road Design using Bilevel
Programming and Greedy–Seeded Simulated Annealing: A Case Study of Kinshasa}

\author[matangany@tut.ac.za]
{Yves Matanga\fnref{myfootnote}}\cortext[mycorrespondingauthor]{}
\cortext[mycorrespondingauthor]{All data and source code used in this study are available in the KANISA library, currently under development, at the GitHub repository \cite{matanga2025kanisa_transportation}}
\ead{matangany@tut.ac.za}
\ead[url]{www.tut.ac.za}

\author[duc@tut.ac.za]{Chunling Du}
\ead[url]{www.tut.ac.za}

\author[vanwykea@tut.ac.za]{Etienne van Wyk}
\ead[url]{www.tut.ac.za}
\address{Department of Computer Systems Engineering, Tshwane University of Technology, 2 Aubrey Matlakala St, Soshanguve South, South Africa}

\begin{abstract}
\noindent\textbf{Context}: The city of Kinshasa faces severe traffic congestion, requiring strategic infrastructure capacity enhancements.  Although a comprehensive master plan \cite{jica2019kinshasa} has been proposed, its implementation requires substantial financial investment, which remains constrained in the Democratic Republic of the Congo (DRC), an emerging economy. This research proposes a traffic flow–based algorithm to support the development of priority road segments. The objective is to enable more effective prioritisation of road construction projects and facilitate the optimal allocation of limited infrastructure budgets.
\noindent\textbf{Methods}: The study was conducted by formulating a standard transport network design problem (TNDP) that included estimated origin-demand data specific to the city of Kinshasa. Given the high computational nature of the 30-node network design, TNDP-relevant metaheuristics (GA, ACO, PSO, SA, TS, Greedy) were used selectively and hybridised to achieve high-quality, stable solutions. A greedy-search-seeded simulated annealing and Tabu search were devised to achieve the design goals.
\noindent\textbf{Results}: Greedy-Simulated Annealing and Greedy-Tabu search yielded the best travel time reduction and the most stable solutions compared to other solvers, also improving network edge betweenness centrality by nearly a scale of two and a half. 
\noindent\textbf{Conclusions}: Road priorities were proposed, including junctions connecting the Bandundu and Kongo Central entry point to main attraction centres (Limete Poids Lourd, Gombe, Airport) and additional inner city areas (Ngaliema, Selembao, Lemba, Masina, Kimwenza).
\end{abstract}
\end{frontmatter}

\section{Introduction}\label{sec1}
\noindent The city of Kinshasa experiences severe congestion that necessitates substantial infrastructure capacity enhancements \cite{jica2019kinshasa, kayisu2024system}. Most of its traffic issues are caused by a lack of roads and road capacity,  poor driving habits, and poor traffic management \cite{kayisu2024system, munga2023dynamic}. The development of a full-blown infrastructural implementation proposed by \cite{jica2019kinshasa} may involve a high-budget initiative within the constrained operational landscape of the emerging DRC. This justifies the need for a road prioritisation scheme based on citywide origin-demand data, considering policymakers' budget constraints.

\par Optimal TNDPs have been formulated in road construction engineering, solved by metaheuristics and approximation methods \cite{leblanc1975algorithm,zhang2009bilevel, iliopoulou2019metaheuristics} given the intractability of exact solutions. Several metaheuristics have been used in the past to solve TNDPs, including genetic algorithms \cite{yin2000genetic,sun2016continuous}, ant colony optimisation \cite{ghaffari2021risk}, particle swarm optimisation \cite{babazadeh2011application, barahimi2021multi}, and simulated annealing \cite{rashidi2016optimal}\cite{chau2025systematic}, with growing attempts to use exact branch and bound methods \cite{rey2020computational}. From a policymaker's point of view, solutions proposed by intelligent algorithms must be repeatable despite the stochastic nature of metaheuristic computation and exhibit an incremental edge-generation process that accommodates policymakers' limited budget allocations. This design philosophy has guided the current research. An optimal road-augmentation solution for the Kinshasa network was thus proposed using a hybrid greedy-simulated annealing and greedy tabu search algorithm. It combines the incremental edge addition of greedy search \cite{medya2018group} with the exploratory local perturbations of simulated annealing \cite{nikolaev2010simulated} or Tabu search \cite{glover2013tabu} to yield high-quality, long-horizon-aware, stable solutions and recommendations.

\par Computational experiments were conducted, including common metaheuristics in TNDP and combinatorial optimisation: GA, ACO, PSO, GrA, SA, and TS, in which Greedy-Simulated Annealing yielded the highest reduction in travel time and solution stability, followed by Greedy-Tabu search and the remaining algorithms, therefore solidifying the usage of greedy-local searches as viable solutions for TNDP. Policy recommendations are proposed for the city of Kinshasa based on the best results obtained from Greedy-Simulated Annealing. All data and code used in the study are available in \cite{matanga2025kanisa_transportation} to support additional computational experiments and modifications to network or origin-demand data. To the best of our knowledge, this study is the first publicly available optimisation and computation examination of network augmentation for the city of Kinshasa. The contributions of the research are:
\begin{enumerate}
    \item An aggregated, analytical, and traffic-informed network analysis of the city of Kinshasa is constructed.
    
    \item A bilevel programming formulation for travel time minimisation is presented for the city of Kinshasa network design problem.

    \item A stable greedy-local simulated annealing and tabu search computational solution is proposed, yielding improved cost optimisation against standard metaheuristics.

    \item Policy Recommendations for priority congestion minimisation in the city of Kinshasa are provided.
\end{enumerate}

\noindent The structure of this paper is as follows. Section 1 introduces the congestion problem context of the city of Kinshasa; Section 2 provides a traffic analysis of the Kinshasa main road network, and Section 3 formulates the optimisation road-relief problem. Section 4 describes the different algorithmic solvers for the TNDP; Section 5 describes the computational experiment settings for the problem at hand; Section 6 presents the results of the computational experiment; Section 7 discusses the findings and provides policy recommendations; and Section 8 concludes the research study.

\section{Traffic Network Modelling and Analysis}

\begin{figure}[H]
    \centering
    \includegraphics[width=1.0\linewidth]{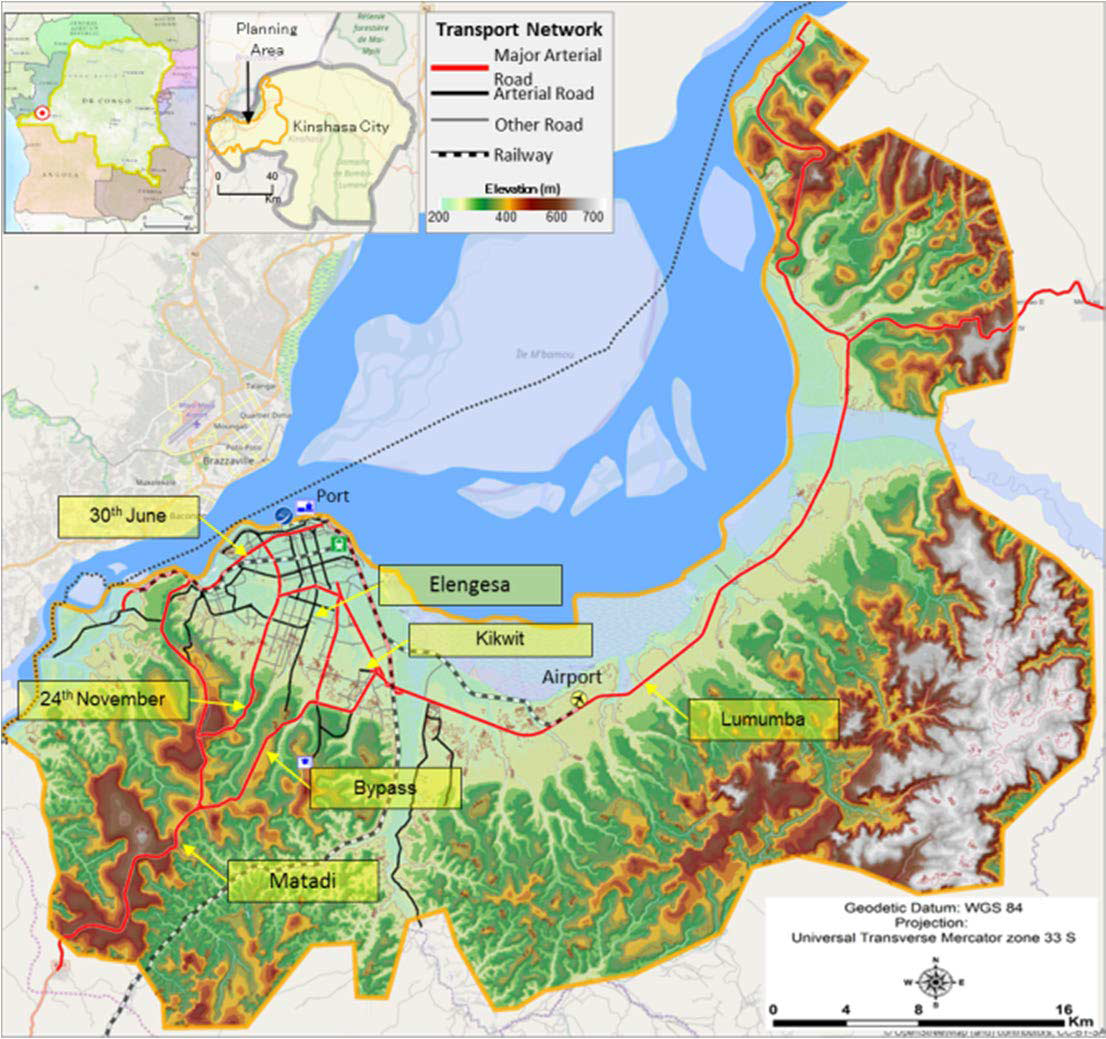}
    \caption{Kinshasa Transport Network - Major Arterial Roads \cite{jica2019kinshasa}}
    \label{fig:kin_city}
\end{figure}

\noindent Kinshasa is the largest city in the Democratic Republic of Congo, with over 16 million people and a surface area of nearly 10,000 km$^2$. The city is most densely populated in its north-western sector, with the majority of its land area still sparsely populated. Figure \ref{fig:kin_city} shows the main arterial roads of the city that are dramatically insufficient in view of its demographic size. 

\begin{figure}[h]
    \centering
 \scalebox{0.9}{\includegraphics[width=1.0\linewidth]{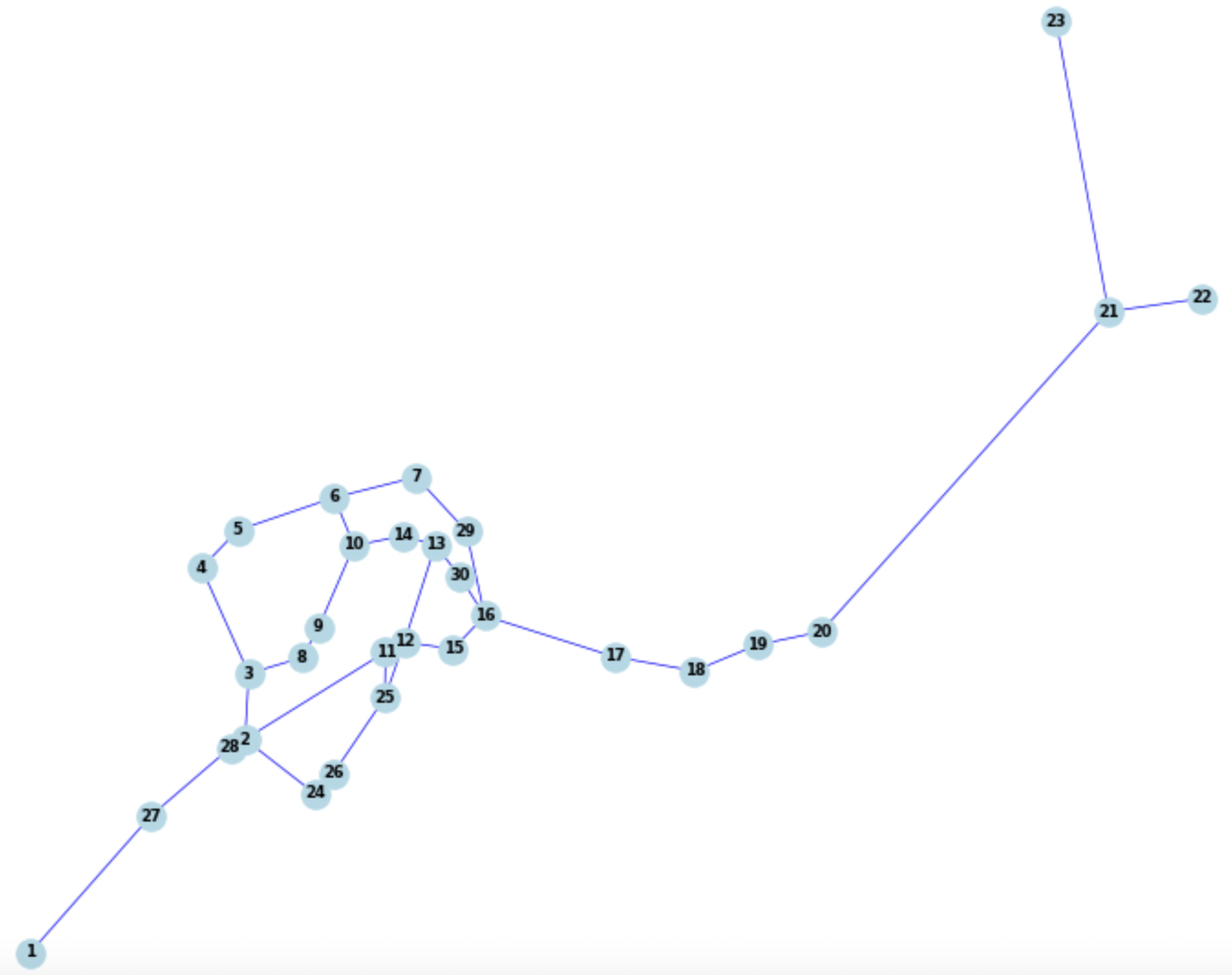}
 }
    \caption{Graph representation of Kinshasa Main Arteries}
    \label{fig:kin_graph_network}
\end{figure}

\noindent The city's main arterial road network can be simplified into a 30-node graph (See Figure \ref{fig:kin_graph_network}) that helps us pinpoint its structural properties and limitations.

\begin{table}[H]
\centering
\caption{Node Description}
\label{tab:results}
\scalebox{0.7}{
\begin{tabular}{|r|rrl|}
\hline
\textbf{Node} & \textbf{Lat} & \textbf{Long} & \textbf{Description} \\
\hline
1  & -4.559058 & 15.174809 & Route de Matadi (Entrée Kinshasa) \\
2  & -4.442848 & 15.255100 & Triangle Matadi Kibala \\
3  & -4.406924 & 15.256513 & UPN \\
4  & -4.349325 & 15.238810 & Av. de l'École – Binza \\
5  & -4.328773 & 15.252333 & Mont Ngaliema \\
6  & -4.310907 & 15.288488 & Bd du 30 Juin \\
7  & -4.299709 & 15.319240 & Gare Centrale \\
8  & -4.398246 & 15.276497 & Selembao (Auto Stop) \\
9  & -4.381511 & 15.282728 & Sanatorium \\
10 & -4.337062 & 15.295951 & Pierre Mulele \\
11 & -4.395099 & 15.307741 & Triangle Campus \\
12 & -4.389760 & 15.314763 & Rond Point Ngaba \\
13 & -4.336819 & 15.326397 & Av. de l'Université \\
14 & -4.331834 & 15.314427 & Bd Triomphal \\
15 & -4.393572 & 15.333011 & Lemba \\
16 & -4.375441 & 15.344869 & Échangeur 1 \\
17 & -4.397829 & 15.393514 & Masina \\
18 & -4.405498 & 15.423430 & Av. Ndjoku \\
19 & -4.391380 & 15.446782 & Aéroport Ndjili \\
20 & -4.384103 & 15.470908 & Nsele \\
21 & -4.209434 & 15.578420 & RP Nsele \\
22 & -4.202280 & 15.613246 & Menkao \\
23 & -4.051282 & 15.558907 & Maluku \\
24 & -4.472585 & 15.281341 & Arrêt Gare \\
25 & -4.420154 & 15.307442 & UNIKIN \\
26 & -4.461385 & 15.288240 & Kimwenza 2 \\
27 & -4.484759 & 15.219752 & Benseke \\
28 & -4.447334 & 15.249644 & Wenze Matadi Kibala \\
29 & -4.329183 & 15.337959 & Limete PL \\
30 & -4.353936 & 15.335540 & Limete R \\
\hline
\end{tabular}
}
\end{table}

\begin{figure}[H]
    \centering
    \includegraphics[width=1.0\linewidth]{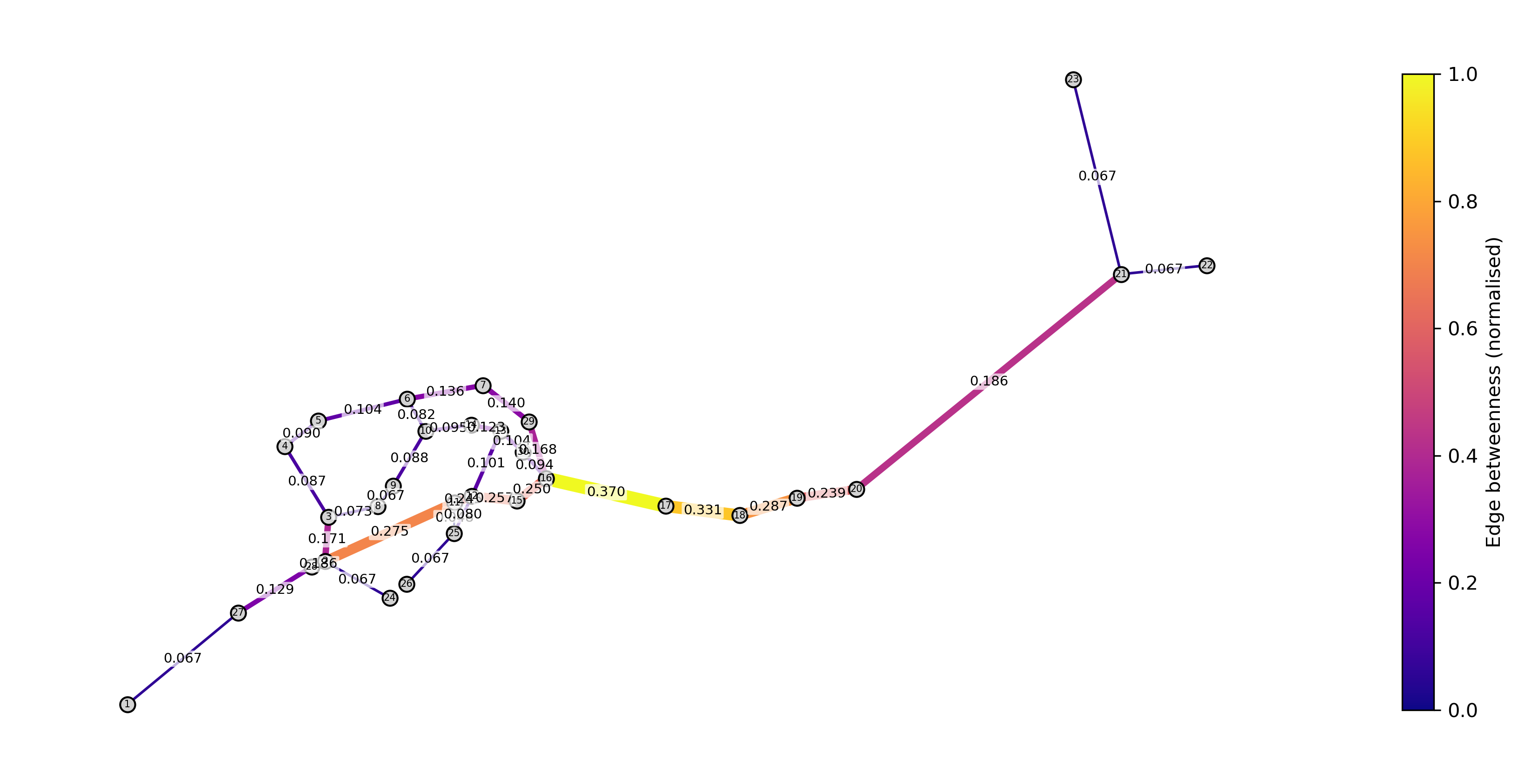}
    \caption{Edge Betweenness centrality - Kinshasa Traffic Network}
    \label{fig:kin_edge_betweeness}
\end{figure}

\noindent One standard indicator of structural bottlenecks in a network is edge (or node) betweenness centrality, which assesses the structural load on road junctions. It assigns a higher load score to junctions that most shortest paths must traverse to reach their destinations. The edge betweenness centrality $C_B(e)$ assesses the proportion of shortest path connectivities that pass through an edge $e$ in a given graph:

\begin{equation}
C_B(e,G) = \sum_{e \neq (j,k)} \frac{n_{jk}(e)}{n_{jk}}
\end{equation}
\noindent where $n_{jk}$ is the number of shortest paths between $j$ and $k$, and  $n_{jk}(e)$ the number of $n_{jk}$ that passes through $e$.  From a structural perspective, Figure \ref{fig:kin_edge_betweeness} computes the edge betweenness centrality in the Kinshasa main road network.

\noindent This data shows that the main structural bottlenecks are in the junctions connecting the Bandundu province entry, the airport, and the axis Roint Point Ngaba, Matadi-Kibala-UPN. 

\begin{figure}[H]
    \centering
    \scalebox{0.7}{
    \includegraphics[width=0.9\linewidth]{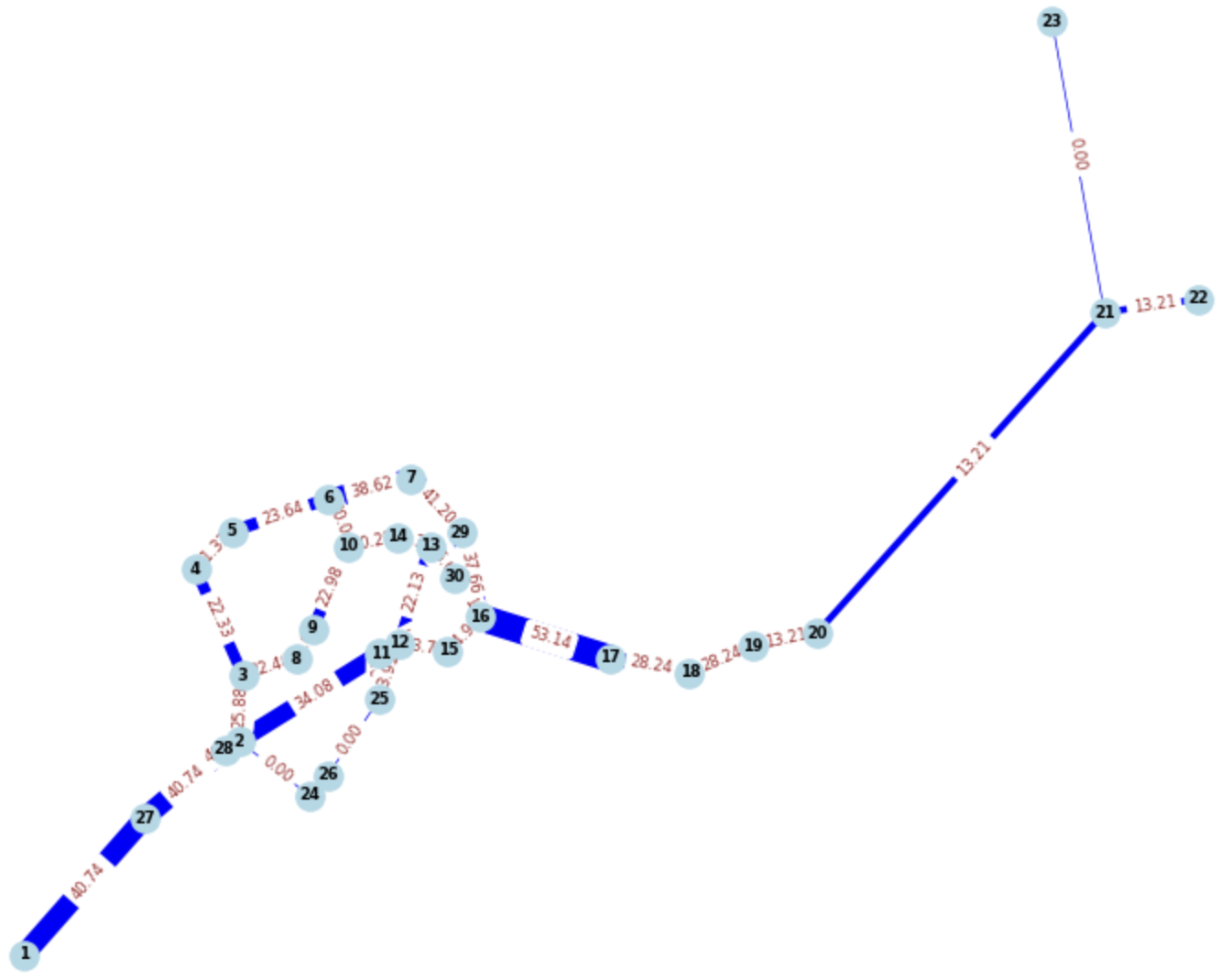}
    }
    \caption{Demand-based Traffic data in the Kinshasa Network as per O-D pair data in the Appendix and flow-based simulation in the lower level problem (Equation \ref{eq:lvl_prob})}
    \label{fig:kin_flow_demand}
\end{figure}

\noindent In the same vein, traffic flow estimated from the Origin-Destination demand data (See Table \ref{tab:normalised_OD_pairs} and Appendix A) \cite{jica2019kinshasa} show high congestion on junctions connecting Masina-Echangeur 1 to the airport, Gombe CBD to the industrial Limete zone, Matadi Kibala entry point heading to the UPN and Ngaba axes, the second CBD road between the 24$^{th}$ November road exiting to Limete and the first CBD road entering Gombe (See Table \ref{tab:g_edges} and Figure \ref{fig:kin_flow_demand}).

\begin{table}[H]
    \centering
    \scalebox{0.7}{
    \begin{tabular}{|c|c|l|c|l|c|c|}
    \hline
    \textbf{Rank} & \textbf{Node 1} & \textbf{Description} & \textbf{Node 2} & \textbf{Description} & \textbf{Betweenness} & \textbf{Traffic Volume} \\
    \hline
    
1  & 16 & Échangeur 1                     & 17 & Masina                      & 0.370115 & 53.144000 \\
2  & 7  & Gare Centrale                   & 29 & Limete Poids Lourd          & 0.140230 & 41.197938 \\
3  & 1  & Route de Matadi (border Kinshasa) & 27 & Benseke                    & 0.066667 & 40.740000 \\
4  & 27 & Benseke                         & 28 & Wenze Matadi Kibala         & 0.128736 & 40.740000 \\
5  & 2  & Triangle Matadi Kibala          & 28 & Wenze Matadi Kibala         & 0.186207 & 40.740000 \\
6  & 10 & Pierre Mulele                    & 14 & Bd Triomphal                & 0.095402 & 40.218984 \\
7  & 6  & Bd du 30 Juin                    & 7  & Gare Centrale               & 0.135632 & 38.624990 \\
8  & 16 & Échangeur 1                      & 29 & Limete Poids Lourd          & 0.167816 & 37.663009 \\
9  & 13 & Av. de l'Université              & 30 & Limete Résidentiel          & 0.104215 & 36.609607 \\
10 & 15 & Lemba                             & 16 & Échangeur 1                 & 0.249808 & 34.976453 \\
11 & 13 & Av. de l'Université              & 14 & Bd Triomphal                & 0.122989 & 34.334984 \\
12 & 2  & Triangle Matadi Kibala          & 11 & Triangle Campus             & 0.274713 & 34.075075 \\
13 & 6  & Bd du 30 Juin                    & 10 & Pierre Mulele               & 0.081609 & 30.092984 \\
14 & 18 & Av. Ndjoku                       & 19 & Aéroport Ndjili             & 0.287356 & 28.244000 \\
15 & 17 & Masina                            & 18 & Av. Ndjoku                  & 0.331034 & 28.244000 \\
16 & 16 & Échangeur 1                      & 30 & Limete Résidentiel          & 0.094253 & 26.174091 \\
17 & 2  & Triangle Matadi Kibala          & 3  & UPN                         & 0.171264 & 25.876923 \\
18 & 12 & Rond Point Ngaba                 & 15 & Lemba                       & 0.256705 & 23.792452 \\
19 & 5  & Mont Ngaliema                    & 6  & Bd du 30 Juin               & 0.103831 & 23.636066 \\
20 & 9  & Sanatorium                        & 10 & Pierre Mulele               & 0.088123 & 22.984857 \\
21 & 8  & Selembao (Auto Stop)             & 9  & Sanatorium                  & 0.067433 & 22.602000 \\
22 & 3  & UPN                               & 8  & Selembao (Auto Stop)        & 0.072797 & 22.401999 \\
23 & 3  & UPN                               & 4  & Av. de l'École – Binza      & 0.086973 & 22.332005 \\
24 & 12 & Rond Point Ngaba                 & 13 & Av. de l'Université         & 0.100766 & 22.133974 \\
25 & 4  & Av. de l'École – Binza           & 5  & Mont Ngaliema               & 0.090038 & 21.316005 \\
26 & 11 & Triangle Campus                  & 12 & Rond Point Ngaba            & 0.244828 & 20.160798 \\
27 & 12 & Rond Point Ngaba                 & 25 & UNIKIN                      & 0.080460 & 13.914277 \\
28 & 11 & Triangle Campus                  & 25 & UNIKIN                      & 0.048276 & 13.914277 \\
29 & 20 & Nsele                             & 21 & RP Nsele                    & 0.186207 & 13.212000 \\
30 & 19 & Aéroport Ndjili                  & 20 & Nsele                       & 0.239080 & 13.212000 \\
31 & 21 & RP Nsele                         & 22 & Menkao                      & 0.066667 & 13.212000 \\
32 & 2  & Triangle Matadi Kibala          & 24 & Arrêt Gare                  & 0.066667 & 0.000000 \\
33 & 25 & UNIKIN                            & 26 & Kimwenza 2                  & 0.066667 & 0.000000 \\
34 & 21 & RP Nsele                         & 23 & Maluku                      & 0.066667 & -0.000000 \\

\hline
    \end{tabular}
}
    \caption{Traffic volume and Edge Betweenness Centrality in the Kinshasa Road Network}
    \label{tab:g_edges}
\end{table}

\begin{figure}[H]
    \centering
    \includegraphics[width=0.8\linewidth]{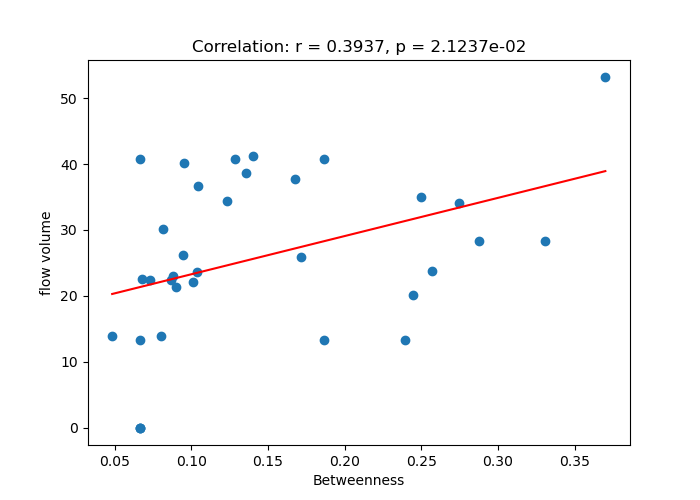}
    \caption{Correlation between network structural edge betweenness centrality and estimated traffic flow volume}
    \label{fig:corr}
\end{figure}

\noindent Correlation analysis in Figure \ref{fig:corr} pinpoints the inherent relationship between network structure and traffic flow, reinforcing the need for network augmentation. Section \ref{sec:prob_formulation} presents the optimisation formulation for network augmentation that accounts for traffic origin-destination pair demand data, user road-choice behaviour, and budget constraints.

\section{Problem formulation}
\label{sec:prob_formulation}

\begin{figure}[H]
    \centering
    \includegraphics[width=1.0\linewidth]{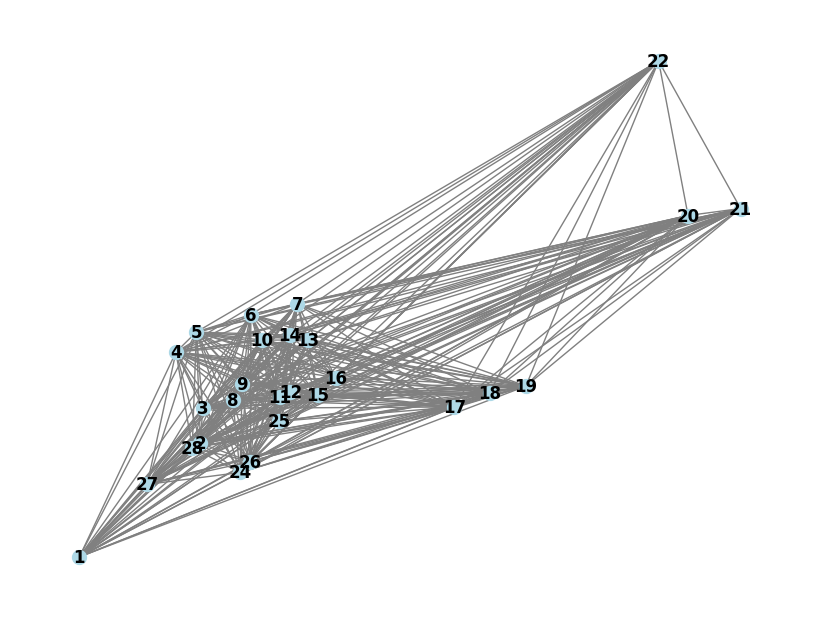}
    \caption{Complete graph $G_c$ of the initial network $G_0$ with 456 edges}
    \label{fig:comp_g}
\end{figure}

\noindent TNDPs revolve around augmenting a network to maximise or minimise a cost function related to travel time, network centrality, or other metrics \cite{gupta2011approximation,medya2018group,jia2019review}. We consider a multi-node graph edge augmentation scenario (See Figure \ref{fig:comp_g}).

\noindent Binary decision variables $y_{ij}$ are denoted from the complete graph originating from the initial network, stating whether a given edge $(i,j)$ should be added to the initial network. Let $G^a$, be a set of all missing edges from $G_0$:
\begin{equation}
    y_{i,j}=\begin{cases}
        1 \text{ if edge }i,j \in G^a\text{is added to }G_0\\
        0 \text{ otherwise}
    \end{cases}
\end{equation}

\noindent A TNDP is typically formulated as a bilevel optimisation problem \cite{leblanc1975algorithm, madadi2024hybrid, yan2024study}. The upper-level problem (ULP) aims to find the optimal $y^*$ vector that minimises travel time (or any other quality cost). In contrast, the lower-level problem (LLP) dynamically integrates network traffic flows based on user road-choice behaviour as the network structure changes.

\begin{align}
  \textbf{ULP}: \underset{y}{Minimise} \text{ }\quad  & \underset{}{T(y) = \sum_{(i,j)\in \Gamma \cup E}x^*_{i,j}t_{ij}(x^*_{i,j})}+q\lambda I(y)\\
   s.t\quad  & \sum_{}^{}c_{i,j}y_{i,j } \leq  B\\
   \quad & y_{i,j}\in\{0,1\}\\
   \quad & q\in \{0,1\}\\
   \quad & x^*_{i,j} = \text{arg min }F(x,f,y)\\
   \textbf{LLP}: \underset{x, f}{Minimise} \text{ }\quad & \underset{}{F(x,f,y) = \sum_{(i,j)\in \Gamma\cup E} \int_{0}^{x_{i,j}}t_{i,j}(w)dw} \label{eq:lvl_prob}\\
   s.t\quad & x_{i,j}=\sum_{P_{r,s}\in E\cup \Gamma}\sum_{k\in P_{r,s}}f_k\delta_k^{i,j},\forall (i,j) \in E\cup \Gamma\\
   \quad & \sum_{k\in P_{r,s}}f_k=d_{r,s},\forall k\in P_{r,s}\\
   \quad & f_k\geq 0\\
   \quad & x_{i,j}\geq 0\\
   \quad & \delta_k^{i,j}\in \{0,1\}
\end{align}
\noindent where $y_{i,j}$ is a decision variable to add or remove an edge $(i,j)$, $x_{i,j}$ is the overall flow rate of the edge $(i,j)$, $f_k$ is the flow of a path $k$ in a given origin destination pair $(r,s)$ and $\delta_k^{i,j}$ a network structure variable that excludes a flow $f_k$ if it does not go through $x_{i,j}$. $d_{r,s}$ origin-destination demand data in selected node-pairs, $c_{ij}$ is the budget cost per edge, $q$ is a design decision to penalise solutions with edge intersections (i.e $I(.)$), $\gamma$ is a penalty factor $I(.)$. The travel time per edge is typically a variant of the BPR formula proposed by the US Bureau of Public Roads in 1964, and widely used \cite{babazadeh2011application,koh2007solving,zhang2009bilevel}

\begin{equation}
    \text{BPR Function } t = t_0(1+\alpha(\frac{V}{C})^4) 
\end{equation}

\noindent where $V$ is the traffic volume, $C$ is the road capacity, $t_0$ is the free flow time, and $\alpha$, a constant typically set to 0.15 in literature \cite{jica2019kinshasa}.
\noindent We set $x_{ij}=\frac{V_{i,j}}{C_{ij}}$ and simplify  $t_0$ as a correlate of road length in our study.

\begin{align}
t(x) = t_{ij}(x_{ij}) = d_{ij}(1+\alpha x_{ij}^4) 
% T(x,\alpha) = \int_{0}^{x}t(w)dw =  d_{ij}(x_{i,j}>\frac{1}{5}\alpha x_{ij}^5)
\end{align}

\noindent In this study, we consider two designs: an unconstrained design ($q=0$) that connects edges in the network without restrictions on edge intersections, and a constrained design ($q=1$) that penalises networks with edge intersections, thereby simulating a typical natural design pattern. The computational complexity of integer nonlinear TNDPs requires the use of combinatorial approximation methods discussed in section \ref{sec:sol_techniques}.

\section{Solution techniques}\label{sec:sol_techniques}
\noindent The heavy computational burden of TNDPs necessitates the use of approximation methods. In this section, we review the established research algorithms commonly used for TNDP and combinatorial optimisation \cite{chau2025systematic}.

\subsection{Greedy Algorithm}
\noindent Greedy algorithms (GrA) are short-horizon, iterative edge-augmentation algorithms. At each iteration, until budget constraints are reached, the most cost-optimal edge is added to the network, one at a time \cite{medya2018group}. It is a combinatorial algorithm that efficiently provides a decent solution to a large, intractable problem; however, it has the limitation of yielding sub-optimal solutions due to the limited-horizon nature of edge addition. Algorithm \ref{alg:greedy} provides a pseudocode of the procedure.

\begin{algorithm}[H]
\setstretch{0.3}
\SetAlgoLined
  Let $G=(V,E)$ be an initial graph, $\Gamma\equiv$ a set of candidate edges\;
  Let $E_s=\emptyset$, an empty set of candidate solution edges\;
  Let $D(E_s) = \sum_{(i,j)\in E_s}d_{i,j}$\;
  \While{$D(E_s)<d_k$}{
    \For{$e\in \Gamma/E_s$}{
    \If{$d(e) \leq d_k-D(E_s)$}{
        $E_e\leftarrow E \cup \text{ }\{e\}$\;
        $[\hat{x^*_e}]\leftarrow \text{Solve LLP[}E_e\text{]}$\;
    }
       
    }  
    $e^* = $ arg min $ULP[E_e,\hat{x^*_e}]$\;
    $E\leftarrow E \cup \text{ }\{e^*\}$\;
    $E_s\leftarrow E_s \cup \text{ }\{e^*\}$\;
    $D(E_s) = \sum_{(i,j)\in E_s}d_{i,j}$\;
  }
  return $E_s$\;
 \caption{Greedy Algorithm}
 \label{alg:greedy}
\end{algorithm}

\noindent \newline At each iteration, a sample candidate edge is selected from $\Gamma/E_s$ and added to a novel network $G_e$ with more edge additions. The edge that yields the best cost function minimisation is added to the candidate solution pool $E_s$, and, as a result, the network and the candidate subset are updated, along with the termination constraint. This process continues until a termination criterion is reached. An additional benefit of greedy algorithms is their short-horizon edge prioritisation scheme, which selects the most valuable edges within the current horizon, a feature that aligns with our design philosophy.

\subsection{Genetic algorithms}
\noindent Genetic algorithms (GA) are a family of evolutionary algorithms inspired by the theory of evolution, specifically the concept of natural selection \cite{goldberg1989genetic}. Search agents using three main genetic processes evolve generation after generation, via \textbf{elitism}, a selection of the fittest individuals moving to the next generation, \textbf{crossover}, mating of parents to generate better hybrids, and \textbf{mutation}, perturbations of selected agent properties to create new breeds. While elitism preserves good solutions, mutation explores the solution space, and crossover intensifies or fine-tunes them. Algorithm \ref{alg:ga} provides a pseudocode of the procedure.

\begin{algorithm}[H]
\setstretch{0.6}
\SetAlgoLined
  Let $Y_0 = \{\textbf{e}=(e_1,e_2,..,e_p)|\sum_{i=1}^{p}d(e_i) \leq B\}$\;
  Set pop size N, max\_iter $K$ and $k=0$, BUB=$-\infty$\;
  Set elite and crossover count Ne and  Nc\; 
  Randomly generate initial population: $y^0 \in Y_0$\;
  Compute fitness values for each $y_j^0$\;
  Find ($y^*$,BUB) = min(f($y_j^0\in Y_0$),BUB)\;
 \While{$k < K$ and \textbf{heuristic stop} not reached}{
  Add $Ne$ elite vectors to $Y_{k+1}$\;
  \For{j=1 to $\frac{Nc}{2}$}{
    \textbf{Select} two parents $y_p$ and $y_q$ in $Y_k$\;
    [$y_{c_1},y_{c_2}$] = \textbf{cross\_over}($y_q$,$y_p$)\;
    Add $y_{c_1},y_{c_2}$ to $Y_{k+1}$\;
  }
  \For{l=1 to (N-Nc-Ne)}{
     \textbf{Select} a parent $y_p$ in $Y_k$\;
     [$y_{m}$] = \textbf{mutate}($y_p$)\;
     Add $y_{m}$ to $Y_{k+1}$\;
  }
 Find ($y^*$,BUB) = min(f($y_j\in Y_{k+1}$),BUB)\;
  $k=k+1$\;
 }
 \textbf{return} $(y^*,\text{BUB})$\;
 \caption{Typical GA procedure}
 \label{alg:ga}
\end{algorithm}

\subsection{Tabu Search}
\noindent The Tabu search algorithm is a heuristic approach developed by Glover \cite{glover2013tabu} that aims to find better solutions from an initial estimate by directing the search towards novel regions while keeping a record of previously visited domains (the tabu list).  A tabu search algorithm will have the following components \cite{abido2002optimal}: The current solution, moves, the set of candidate moves, and tabu restrictions. The \textbf {current solution} $y_{current}$ is the current main search vector central to generating the neighbourhood of search. \textbf{Moves} refers to the philosophy used to generate trials around $y_{current}$. The \textbf{set of candidate moves} is a group of trial solutions around $y_{current}$. \textbf{Tabu restrictions} refer to the set of conditions to make on moves that prevent reaching forbidden places. A tabu list is thus updated to contain solutions not to be rediscovered. The \textbf{aspiration criterion (level)}  is a rule to override tabu restrictions important to allow recycling back to some regions when relevant (i.e., the objective function improved better than in the previous move).  This adds some flexibility in the tabu search to enhance attractive moves. Algorithm \ref{alg:ts} provides a pseudocode of the procedure.

\begin{algorithm}[H]
\setstretch{0.6}
\SetAlgoLined
  Choose initial solution $y_0$\;
  Set $best \leftarrow y_0$, $f_{best} \leftarrow f(y_0)$\;
  Initialize Tabu list $T \leftarrow \emptyset$\;
  Set max\_iter $K$ and iteration counter $k=0$\;
 \While{$k < K$ and \textbf{heuristic stop} not reached}{
   Generate neighbourhood $N(y_k)$\;
   \For{each $y' \in N(y_k)$}{
      \If{$y' \notin T$ and $f(y') < f(y_k)$}{
         $y_{cand} \leftarrow y'$\;
         $f_{cand} \leftarrow f(y')$\;
      }
   }
   Select best admissible candidate $y_{cand}$\;
   $y_{k+1} \leftarrow y_{cand}$\;
   Update Tabu list $T \leftarrow T \cup \{y_{cand}\}$\;
   If $|T| > tenure$, remove oldest entry from $T$\;
   \If{$f(y_{cand}) < f_{best}$}{
      $best \leftarrow y_{cand}$\;
      $f_{best} \leftarrow f(y_{cand})$\;
   }
   $k \leftarrow k+1$\;
 }
 \textbf{return} $(best,f_{best})$\;
 \caption{Typical Tabu Search procedure}
 \label{alg:ts}
\end{algorithm}

\subsection{Simulated Annealing}
\noindent Simulated annealing is a physics-inspired metaheuristic algorithm that simulates the careful cooling of a metal to reach an equilibrium state with minimal energy, that is, a symmetric alignment of atoms in the matter \cite{delahaye2018simulated}. It belongs to the family of solution improvement methods based on local neighbour generation \cite{aarts2013simulated}, which permits temporary exit from the current good region to escape local optima, with a decreasing tolerance for poorer candidate regions over time. Simulated annealing is typically used for discrete optimisation problems and can also be adapted to continuous optimisation \cite{aarts2013simulated}. Algorithm \ref{alg:sa} provides the pseudocode of the procedure.

\begin{algorithm}[H]
\setstretch{0.6}
\SetAlgoLined
  Choose initial solution $y_0$\;
  Evaluate $f(y_0)$ and set $best \leftarrow x_0$, $f_{best} \leftarrow f(y_0)$\;
  Set initial temperature $T \leftarrow T_0$, cooling rate $\alpha$, and minimum temperature $T_{\min}$\;
  Set maximum iterations $K$ and iteration counter $k \leftarrow 0$\;
 \While{$k < K$ \textbf{and} $T > T_{\min}$}{
   Generate a random neighbour $y'$ of $y_k$\;
   Evaluate $\Delta f \leftarrow f(y') - f(y_k)$\;
   \eIf{$\Delta f < 0$}{
      Accept new solution: $y_{k+1} \leftarrow y'$\;
   }{
      Compute acceptance probability $P \leftarrow e^{-\Delta f / T}$\;
      Generate random number $r \sim U(0,1)$\;
      \If{$r < P$}{
         Accept worse solution: $y_{k+1} \leftarrow y'$\;
      }
      \Else{
         Reject: $y_{k+1} \leftarrow x_k$\;
      }
   }
   \If{$f(y_{k+1}) < f_{best}$}{
      $best \leftarrow y_{k+1}$\;
      $f_{best} \leftarrow f(y_{k+1})$\;
   }
   Update temperature: $T \leftarrow \alpha \times T$\;
   $k \leftarrow k + 1$\;
 }
 \textbf{return} $(best, f_{best})$\;
 \caption{Simulated Annealing Algorithm}
 \label{alg:sa}
\end{algorithm}

\subsection{Particle Swarm Optimisation}
\noindent Particle swarm optimisation is a metaheuristic inspired by the foraging behaviour of birds as they collectively navigate to locate food sources. This search behaviour leverages each bird's local information and the collective intelligence of all birds to perform the search. PSO is generally used for continuous nonconvex optimisation using the following iterative search equations:

\begin{eqnarray}
    v_j^{k+1}=\omega^kv_j^{k}+ c_1r_1^k(p_j^k-y_j^k)+c_2r_2^k(g_{(j)}^k-y_j^k)\\
    y_j^{k+1} = y_j^k + v_j^{k+1}
\end{eqnarray}

\noindent where $v_j^{k+1}$ is the search direction of a given particle for the next iteration, which is a function of its current velocity $v_j^k$, its best location thus far \textbf{$p_j^k$} (cognitive learning) and the best location $g_{(j)}^k$ (social learning) in the neighbourhood of the particle ($g_j^k$) or within the whole swarm ($g^k$). The parameters $c_1$ and $c_2$ represent acceleration parameters for the cognitive and social learning components. $r_1^k,r_2^k\in [0,1]$ are uniformly randomly generated numbers that simulate the stochastic behaviour of the swarm. The inertial parameter $w^k$ defines how willing a given particle is to maintain its current direction. Together with $c_r$ and $s_r$, they determine the bias towards exploration and exploitation. The higher the inertial parameter, the more exploratory the search. This iterative process continues until the maximum number of iterations is reached or a heuristic stopping criterion terminates the search. Given the combinatorial nature of the problem, binary swarm optimisation proposed by \cite{kennedy1997discrete} is used in this work. Both binary and continuous PSO follow the same nature, with the difference that in the binary version, the velocity and particle values are restricted to the range [0,1]:

\begin{eqnarray}
    v'_{k+1} = sig(v_{k+1}) = \frac{1}{1+e^{-v_{k+1}}}\\
    x_{k+1} = \begin{cases}1 \text{ if } sig(v'_{k+1}) > r\\
    0\text{ otherwise}
    \end{cases}
\end{eqnarray}

\noindent where $r$ is a random variable between 0 and 1. Algorithm \ref{alg:bpso} provides a pseudocode of the binary PSO procedure.

\begin{algorithm}[H]
\setstretch{0.2}
\SetAlgoLined
  Let $X_0=[x^l, x^u],V=[-v_{max},v_{max}]$\;
  Set swarm size N, max\_iter $K$ and $k=0$\;
  Randomly generate initial population: $y_j  \in X_0$\;
  Randomly generate initial velocities: $ v_j \in V$\;
  Set $p_j^k$ for every particle to $y_j^k$\;
  Compute the swarm initial best point ($g^k$, BUB)\;
 \While{$k < K$ and \textbf{heuristic stop} not reached}{
  \For{j=1:N}{
    $v_j^{k+1}=\omega^kv_j^{k}+r_1^kc_1(p_j^k-x_j^k)+r_2^kc_2(g^k-x_j^k)$\;
    $v'_{k+1} = sig(v_{k+1})$\;
    $y_{k+1} = \begin{cases}1 \text{ if } sig(v'_{k+1}) > r\\
    0\text{ otherwise}
    \end{cases}$\;
    $f_j^{k+1} = f(y_j^{k+1})$\;
    $p_j^k = \text{arg min}(\{f_j^{k+1},f_j^k\})$\;
    $g^k = \text{arg min}(\{f(p_j^k),$BUB$\})$\; 
  }
  $k=k+1$\;
 }
 \textbf{return} $(y^*=g^k,\text{BUB})$\;
 \caption{Typical Binary PSO procedure}
 \label{alg:bpso}
\end{algorithm}

\subsection{Ant Colony Optimisation}
\noindent Ant Colony optimisation is a metaheuristic inspired by the food search of real ants, operating in a multipath solution space, in which ants indirectly communicate by leaving pheromone trails on segments to signal the desirability level of junctions. These pheromone deposits influence the global behaviour of ants towards promising paths or solutions. ACO is well-suited for combinatorial optimisation problems \cite{dorigo2018ant} and commonly used in TNDPs \cite{ghaffari2021risk}.

\noindent The algorithmic process of an ACO mechanism goes as follows. Initially, $m$ ants are selected, and all possible network edges are assigned an initial pheromone level $\tau_{i}=\tau_0$. At each iteration, each ant constructs a complete path to the problem within feasible constraints (i.e., Budget). During the path construction process, a random starting node is selected, and the complete path is generated edge by edge from a possibility pool, based on transition-pheromone-informed probability scores $P_{ij}$. Each complete path is then evaluated on the cost function, and each edge pheromone level is updated, respectively, on account of its contributions to good paths. This process continues until the algorithm reaches a maximum count or a heuristic stop criterion is met. The transition probability score is defined as follows
\begin{eqnarray}
    P_{ij} = P(j|i) =  \frac{[\tau_{ij}]^\alpha [\eta_{ij}]^\beta}
            {\sum\limits_{l \in N_i} [\tau_{il}]^\alpha [\eta_{il}]^\beta}
\end{eqnarray}

\noindent where $\tau_{ij}$ is the edge pheromone level, $\eta_{ij}$ is the edge heuristic information and $N_i$ is the feasible neighbourhood. The coefficients $\alpha$ and $\beta$ describe, respectively, the dominance ratio of the pheremone score over the heuristic information or vice versa. The heuristic role of the design $\eta$ is defined for the type of problem in question. The update rule of the pherenomone score $\tau_{ij}$ is given as follows:

\begin{eqnarray}
    \tau_{ij} = (1-\rho)\tau_{ij}>\sum_{a=1}^{m}\Delta\tau_{ij}^a\\
    \Delta\tau_{ij}^a = 
        \begin{cases}
        Q / f_a, & \text{if ant } a \text{ used edge } (i,j)\\
        0, & \text{otherwise}
        \end{cases}
\end{eqnarray}

\noindent where $\rho$ is the evaporation rate,  $\Delta\tau_{ij}^a$ is the pheremone contribution of each ant to the edge.

\begin{algorithm}[H]
\setstretch{0.6}
\SetAlgoLined
  Initialise pheromone levels $\tau_{ij} \leftarrow \tau_0$ for all edges $(i,j)$\;
  Set number of ants $m$, evaporation rate $\rho$, and maximum iterations $K$\;
  Initialise iteration counter $k \leftarrow 0$\;
  \While{$k < K$}{
    \For{each ant $a = 1$ \KwTo $m$}{
        Place ant $a$ on a random starting node\;
        \While{solution of ant $a$ not complete}{
            Select next node $j$ from current node $i$ using transition rule $P_{ij}$\;
            Move to node $j$ and add edge $(i,j)$ to ant’s solution\;
        }
        Evaluate objective function $f_a$ for constructed solution\;
    }
    \textbf{Pheromone update:}\;
    \For{each edge $(i,j)$}{
        Update $\tau_{ij}$\;
    }
    Find best solution $best_k = \arg\min(f_a)$ among all ants\;
    \If{$f(best_k) < f_{best}$}{
        $best \leftarrow best_k$\;
        $f_{best} \leftarrow f(best_k)$\;
    }
    $k \leftarrow k + 1$\;
  }
  \textbf{return} $(best, f_{best})$\;
  \caption{Ant Colony Optimisation (ACO) Algorithm}
  \label{alg:aco}
\end{algorithm}
\subsection{Greedy-Seeded Local searches}
\noindent Rigorous exact methods for discrete network design problems offer the benefits of obtaining solutions of the highest quality, deterministically repeatable run after run, traits that are viable for policy makers in infrastructure development \cite{rey2020computational, neumaier2004complete}. However, given the heavy computational nature and intractability of exact methods for moderate-to-large node graphs, meta-heuristics are the most efficient solution approaches. Nevertheless, a good meta-heuristic solution should be computationally efficient, moderately stable, and of high quality. Given this design philosophy, we propose combining the deterministic, efficient greedy search with exploratory local search algorithms. The rationale is that a greedy search would yield an efficient, repeatable solution, which local perturbation algorithms could further improve by exploring edge neighbourhoods or complete edge mutations. This algorithm approach, by virtue of its traits, would yield efficient, stable, and high-quality solutions. Hybrid greedy-simulated-annealing(Gr-SA) and greedy-tabu-search (Gr-TS) algorithms were proposed for their exploratory capabilities.

\section{Computational Experiment}
\subsection{Datasets and Experimental Settings}
\noindent In the current study, the TNDP problem described in section \ref{sec:prob_formulation} was solved using eight optimisation methods: Genetic algorithms (GA), Greedy algorithm (GrA), Simulated Annealing (SA), Ant Colony optimisation (ACO), Tabu Search (TS), particle swarm optimisation (PSO), the Greedy-simulated annealing hybrid (Gr-SA) and the Greedy-Tabu-Search hybrid (Gr-TS). Table \ref{tab:meta} describes each algorithm setting.

\begin{table}[h]
    \centering
    \scalebox{0.7}{
    \begin{tabular}{|c|l|}
        \hline
         GA &\begin{tabular}{l}crossover: single point\\ mutation: random\\selection: stochastic uniform \\
         \end{tabular}\\
         \hline
         GrA & type: forward sequential addition\\
          \hline
         SA & $T_0=100, T_{min}=1e-3, \alpha=0.97 $ \\
         \hline
         SA-GS & $T_0=1, T_{min}=1e-3, \alpha=0.97$ \\
         \hline
         ACO & $ Q=100, \rho = 0.5, \alpha=2.0,\beta=2.0$ \\
         \hline
         TS & \begin{tabular}{l}
         neighSize = 20\\
         tenu\_iter=5\\
         \end{tabular}\\
         \hline
         TS-GS & \begin{tabular}{l}
         neighSize = 20\\
         tenu\_iter=5\\
  
         \end{tabular}\\
          \hline
         PSO & $c_1=c_2=2.0, w_{max}=0.9,w_{min}=0.3,v_{max}=4.0$ \\
          \hline
          all & max\_pop\_size = 20, max\_iter = 200 (i.e. Fev = 4000)\\
          \hline
    \end{tabular}
    }
    \caption{Parameter configuration of test algorithms}
    \label{tab:meta}
\end{table}

\noindent A network design budget of 100 Km was used in the study. The Lower Level Problem was solved using the  IPOPT local NLP solver \cite{IPOPT} within the Python Pyomo optimisation modelling library \cite{bynum2021pyomo}. Origin-Demand pair data (i.e. 78 non-zero data) were estimated from the JICA survey analysis (See Table \ref{tab:normalised_OD_pairs} and Appendix A) \cite{jica2019kinshasa}. The benchmarking criteria for the algorithms are described in Section \ref{sec:bench}. Computational experiments were conducted on an 8GB M1 MacBook Air.

\begin{table}[H]
\centering
\caption{Normalised Matrix of demands between OD pairs}
% \caption{Normalised Matrix of demands between OD pairs (in thousand vehicles per hour).}

\scalebox{0.5}{
\begin{tabular}{|c*{30}{c}|}
\hline
\textbf{Node} & \textbf{1} & \textbf{2} & \textbf{3} & \textbf{4} & \textbf{5} & \textbf{6} & \textbf{7} & \textbf{8} & \textbf{9} & \textbf{10} & \textbf{11} & \textbf{12} & \textbf{13} & \textbf{14} & \textbf{15} & \textbf{16} & \textbf{17} & \textbf{18} & \textbf{19} & \textbf{20} & \textbf{21} & \textbf{22} & \textbf{23} & \textbf{24} & \textbf{25} & \textbf{26} & \textbf{27} & \textbf{28} & \textbf{29} & \textbf{30} \\
\hline
1 & 0.0 & 1.0 & 0.0 & 0.5 & 1.5 & 0.0 & 2.0 & 0.0 & 0.7 & 2.0 & 0.0 & 2.5 & 4.5 & 3.0 & 0.5 & 1.0 & 1.0 & 0.0 & 0.0 & 0.0 & 0.0 & 0.0 & 0.0 & 0.0 & 0.0 & 0.0 & 0.0 & 0.0 & 0.0 & 0.0 \\
2 & 1.0 & 0.0 & 0.0 & 0.0 & 0.0 & 0.0 & 0.0 & 0.0 & 0.0 & 0.0 & 0.0 & 0.0 & 0.0 & 0.0 & 0.0 & 0.0 & 0.0 & 0.0 & 0.1 & 0.0 & 0.0 & 0.5 & 0.0 & 0.0 & 0.0 & 0.0 & 0.0 & 0.0 & 5.0 & 5.0 \\
3 & 0.0 & 0.0 & 0.0 & 0.0 & 0.0 & 0.0 & 0.0 & 0.0 & 0.0 & 0.0 & 0.0 & 0.0 & 0.0 & 0.0 & 0.0 & 0.0 & 0.0 & 0.0 & 0.0 & 0.0 & 0.0 & 0.0 & 0.0 & 0.0 & 0.0 & 0.0 & 0.0 & 0.0 & 0.0 & 0.0 \\
4 & 0.5 & 0.0 & 0.0 & 0.0 & 0.0 & 0.0 & 0.0 & 0.0 & 0.0 & 0.0 & 0.0 & 0.0 & 0.0 & 0.0 & 0.0 & 0.0 & 0.0 & 0.0 & 0.0 & 0.0 & 0.0 & 0.0 & 0.0 & 0.0 & 0.0 & 0.0 & 0.0 & 0.0 & 0.0 & 0.0 \\
5 & 1.5 & 0.0 & 0.0 & 0.0 & 0.0 & 0.0 & 0.0 & 0.0 & 0.0 & 4.0 & 0.0 & 0.0 & 0.0 & 0.0 & 0.0 & 0.0 & 0.0 & 0.0 & 0.2 & 0.0 & 0.0 & 0.7 & 0.0 & 0.0 & 0.0 & 0.0 & 0.0 & 0.0 & 3.0 & 3.0 \\
6 & 0.0 & 0.0 & 0.0 & 0.0 & 0.0 & 0.0 & 0.0 & 0.0 & 0.0 & 0.0 & 0.0 & 0.0 & 0.0 & 0.0 & 0.0 & 0.0 & 0.0 & 0.0 & 0.0 & 0.0 & 0.0 & 0.0 & 0.0 & 0.0 & 0.0 & 0.0 & 0.0 & 0.0 & 0.0 & 0.0 \\
7 & 2.0 & 0.0 & 0.0 & 0.0 & 0.0 & 0.0 & 0.0 & 0.0 & 0.0 & 0.0 & 0.0 & 0.0 & 0.0 & 0.0 & 0.0 & 0.0 & 4.1 & 0.0 & 0.4 & 0.0 & 0.0 & 0.4 & 0.0 & 0.0 & 0.0 & 0.0 & 0.0 & 0.0 & 7.0 & 7.0 \\
8 & 0.0 & 0.0 & 0.0 & 0.0 & 0.0 & 0.0 & 0.0 & 0.0 & 0.0 & 0.0 & 0.0 & 0.0 & 0.0 & 0.0 & 0.0 & 0.0 & 0.0 & 0.0 & 0.1 & 0.0 & 0.0 & 0.0 & 0.0 & 0.0 & 0.0 & 0.0 & 0.0 & 0.0 & 0.0 & 0.0 \\
9 & 0.7 & 0.0 & 0.0 & 0.0 & 0.0 & 0.0 & 0.0 & 0.0 & 0.0 & 0.0 & 0.0 & 0.0 & 0.0 & 0.0 & 0.0 & 0.0 & 0.0 & 0.0 & 0.0 & 0.0 & 0.0 & 0.0 & 0.0 & 0.0 & 0.0 & 0.0 & 0.0 & 0.0 & 1.0 & 1.0 \\
10 & 2.0 & 0.0 & 0.0 & 0.0 & 4.0 & 0.0 & 0.0 & 0.0 & 0.0 & 0.0 & 0.0 & 0.0 & 0.0 & 0.0 & 0.0 & 0.0 & 0.0 & 0.0 & 0.3 & 0.0 & 0.0 & 0.2 & 0.0 & 0.0 & 0.0 & 0.0 & 0.0 & 0.0 & 3.0 & 3.0 \\
11 & 0.0 & 0.0 & 0.0 & 0.0 & 0.0 & 0.0 & 0.0 & 0.0 & 0.0 & 0.0 & 0.0 & 0.0 & 0.0 & 0.0 & 0.0 & 0.0 & 0.0 & 0.0 & 0.0 & 0.0 & 0.0 & 0.0 & 0.0 & 0.0 & 0.0 & 0.0 & 0.0 & 0.0 & 0.0 & 0.0 \\
12 & 2.5 & 0.0 & 0.0 & 0.0 & 0.0 & 0.0 & 0.0 & 0.0 & 0.0 & 0.0 & 0.0 & 0.0 & 0.0 & 0.0 & 0.0 & 0.0 & 0.0 & 0.0 & 0.1 & 0.0 & 0.0 & 0.0 & 0.0 & 0.0 & 0.0 & 0.0 & 0.0 & 0.0 & 1.0 & 1.0 \\
13 & 4.5 & 0.0 & 0.0 & 0.0 & 0.0 & 0.0 & 0.0 & 0.0 & 0.0 & 0.0 & 0.0 & 0.0 & 0.0 & 0.0 & 0.0 & 0.0 & 0.0 & 0.0 & 0.1 & 0.0 & 0.0 & 0.9 & 0.0 & 0.0 & 0.0 & 0.0 & 0.0 & 0.0 & 0.0 & 0.0 \\
14 & 3.0 & 0.0 & 0.0 & 0.0 & 0.0 & 0.0 & 0.0 & 0.0 & 0.0 & 0.0 & 0.0 & 0.0 & 0.0 & 0.0 & 0.0 & 0.0 & 0.0 & 0.0 & 0.1 & 0.0 & 0.0 & 0.0 & 0.0 & 0.0 & 0.0 & 0.0 & 0.0 & 0.0 & 0.0 & 0.0 \\
15 & 0.5 & 0.0 & 0.0 & 0.0 & 0.0 & 0.0 & 0.0 & 0.0 & 0.0 & 0.0 & 0.0 & 0.0 & 0.0 & 0.0 & 0.0 & 0.0 & 0.0 & 0.0 & 0.1 & 0.0 & 0.0 & 0.0 & 0.0 & 0.0 & 0.0 & 0.0 & 0.0 & 0.0 & 3.0 & 3.0 \\
16 & 1.0 & 0.0 & 0.0 & 0.0 & 0.0 & 0.0 & 0.0 & 0.0 & 0.0 & 0.0 & 0.0 & 0.0 & 0.0 & 0.0 & 0.0 & 0.0 & 0.0 & 0.0 & 0.2 & 0.0 & 0.0 & 1.0 & 0.0 & 0.0 & 0.0 & 0.0 & 0.0 & 0.0 & 0.0 & 0.0 \\
17 & 1.0 & 0.0 & 0.0 & 0.0 & 0.0 & 0.0 & 4.1 & 0.0 & 0.0 & 0.0 & 0.0 & 0.0 & 0.0 & 0.0 & 0.0 & 0.0 & 0.0 & 0.0 & 0.0 & 0.0 & 0.0 & 2.6 & 0.0 & 0.0 & 0.0 & 0.0 & 0.0 & 0.0 & 5.0 & 5.0 \\
18 & 0.0 & 0.0 & 0.0 & 0.0 & 0.0 & 0.0 & 0.0 & 0.0 & 0.0 & 0.0 & 0.0 & 0.0 & 0.0 & 0.0 & 0.0 & 0.0 & 0.0 & 0.0 & 0.0 & 0.0 & 0.0 & 0.0 & 0.0 & 0.0 & 0.0 & 0.0 & 0.0 & 0.0 & 0.0 & 0.0 \\
19 & 0.0 & 0.1 & 0.0 & 0.0 & 0.2 & 0.0 & 0.4 & 0.1 & 0.0 & 0.3 & 0.0 & 0.1 & 0.1 & 0.1 & 0.1 & 0.2 & 0.0 & 0.0 & 0.0 & 0.0 & 0.0 & 0.2 & 0.0 & 0.0 & 0.0 & 0.0 & 0.0 & 0.0 & 3.0 & 3.0 \\
20 & 0.0 & 0.0 & 0.0 & 0.0 & 0.0 & 0.0 & 0.0 & 0.0 & 0.0 & 0.0 & 0.0 & 0.0 & 0.0 & 0.0 & 0.0 & 0.0 & 0.0 & 0.0 & 0.0 & 0.0 & 0.0 & 0.0 & 0.0 & 0.0 & 0.0 & 0.0 & 0.0 & 0.0 & 0.0 & 0.0 \\
21 & 0.0 & 0.0 & 0.0 & 0.0 & 0.0 & 0.0 & 0.0 & 0.0 & 0.0 & 0.0 & 0.0 & 0.0 & 0.0 & 0.0 & 0.0 & 0.0 & 0.0 & 0.0 & 0.0 & 0.0 & 0.0 & 0.0 & 0.0 & 0.0 & 0.0 & 0.0 & 0.0 & 0.0 & 0.0 & 0.0 \\
22 & 0.0 & 0.5 & 0.0 & 0.0 & 0.7 & 0.0 & 0.4 & 0.0 & 0.0 & 0.2 & 0.0 & 0.0 & 0.9 & 0.0 & 0.0 & 1.0 & 2.6 & 0.0 & 0.2 & 0.0 & 0.0 & 0.0 & 0.0 & 0.0 & 0.0 & 0.0 & 0.0 & 0.0 & 0.0 & 0.0 \\
23 & 0.0 & 0.0 & 0.0 & 0.0 & 0.0 & 0.0 & 0.0 & 0.0 & 0.0 & 0.0 & 0.0 & 0.0 & 0.0 & 0.0 & 0.0 & 0.0 & 0.0 & 0.0 & 0.0 & 0.0 & 0.0 & 0.0 & 0.0 & 0.0 & 0.0 & 0.0 & 0.0 & 0.0 & 0.0 & 0.0 \\
24 & 0.0 & 0.0 & 0.0 & 0.0 & 0.0 & 0.0 & 0.0 & 0.0 & 0.0 & 0.0 & 0.0 & 0.0 & 0.0 & 0.0 & 0.0 & 0.0 & 0.0 & 0.0 & 0.0 & 0.0 & 0.0 & 0.0 & 0.0 & 0.0 & 0.0 & 0.0 & 0.0 & 0.0 & 0.0 & 0.0 \\
25 & 0.0 & 0.0 & 0.0 & 0.0 & 0.0 & 0.0 & 0.0 & 0.0 & 0.0 & 0.0 & 0.0 & 0.0 & 0.0 & 0.0 & 0.0 & 0.0 & 0.0 & 0.0 & 0.0 & 0.0 & 0.0 & 0.0 & 0.0 & 0.0 & 0.0 & 0.0 & 0.0 & 0.0 & 0.0 & 0.0 \\
26 & 0.0 & 0.0 & 0.0 & 0.0 & 0.0 & 0.0 & 0.0 & 0.0 & 0.0 & 0.0 & 0.0 & 0.0 & 0.0 & 0.0 & 0.0 & 0.0 & 0.0 & 0.0 & 0.0 & 0.0 & 0.0 & 0.0 & 0.0 & 0.0 & 0.0 & 0.0 & 0.0 & 0.0 & 0.0 & 0.0 \\
27 & 0.0 & 0.0 & 0.0 & 0.0 & 0.0 & 0.0 & 0.0 & 0.0 & 0.0 & 0.0 & 0.0 & 0.0 & 0.0 & 0.0 & 0.0 & 0.0 & 0.0 & 0.0 & 0.0 & 0.0 & 0.0 & 0.0 & 0.0 & 0.0 & 0.0 & 0.0 & 0.0 & 0.0 & 0.0 & 0.0 \\
28 & 0.0 & 0.0 & 0.0 & 0.0 & 0.0 & 0.0 & 0.0 & 0.0 & 0.0 & 0.0 & 0.0 & 0.0 & 0.0 & 0.0 & 0.0 & 0.0 & 0.0 & 0.0 & 0.0 & 0.0 & 0.0 & 0.0 & 0.0 & 0.0 & 0.0 & 0.0 & 0.0 & 0.0 & 0.0 & 0.0 \\
29 & 0.0 & 5.0 & 0.0 & 0.0 & 3.0 & 0.0 & 7.0 & 0.0 & 1.0 & 3.0 & 0.0 & 1.0 & 0.0 & 0.0 & 3.0 & 0.0 & 5.0 & 0.0 & 3.0 & 0.0 & 0.0 & 0.0 & 0.0 & 0.0 & 0.0 & 0.0 & 0.0 & 0.0 & 0.0 & 0.0 \\
30 & 0.0 & 5.0 & 0.0 & 0.0 & 3.0 & 0.0 & 7.0 & 0.0 & 1.0 & 3.0 & 0.0 & 1.0 & 0.0 & 0.0 & 3.0 & 0.0 & 5.0 & 0.0 & 3.0 & 0.0 & 0.0 & 0.0 & 0.0 & 0.0 & 0.0 & 0.0 & 0.0 & 0.0 & 0.0 & 0.0 \\

\hline
\end{tabular}
}
\label{tab:normalised_OD_pairs}
\end{table}
%#2-14 (page 49)
%page (46) 

\noindent Two network design problems are reported in the study, the unconstrained network design ($q=0$), where edges are allowed to cross and the constrained design ($q=1$), where edges are not allowed to cross.

\subsection{Benchmarking criteria}\label{sec:bench}
\noindent To assess the performance of each algorithm in solving the optimisation problem. Each method was tested with an equal number of function evaluations (i.e., Fev = 4000) across 30 optimisation runs. The average objective function value (i.e., network travel time) was used as the primary benchmarking criterion to compare each algorithm,  followed by solution stability, and the average edge betweenness centrality of each new network. The convergence plots and computation time of each algorithm were also reported (See Table \ref{tab:criteria}).

\begin{table}[h]
    \centering
    \begin{tabular}{lll}
    \hline
     N. & Criterion & Acronym.\\
    \hline
         1. & Avg. obj value  & $\tilde{T}(y)$\\
         2. & Obj. n-fold improvement & $\Delta \tilde{T}(y)$ \\
         3. & Avg. network edge between centrality & $\tilde{C}_B$\\
         4. & edge betweenness centrality  n-fold improvement & $\Delta \tilde{C}_B$\\
         5. & Algorithm  solution stability &  $\tilde{S}_m(E_i^m, \textbf{E}^m)$\\
        5. & Convergence plots & - \\
    \hline
    \end{tabular}
    \caption{Assessment criteria}
    \label{tab:criteria}
\end{table}

\noindent The relative improvement in travel time reduction and centrality compared to the original network was also recorded.

\begin{eqnarray}
    \Delta p(i) = \frac{p_0}{p_i}
\end{eqnarray}

\noindent where $p_0$ is the original network performance (i.e., travel time or centrality) and $p_i$ is the network performance after edge additions. A measure of solution stability, $S(E)$, is proposed that assesses how often an algorithm returns the same number of edges across multiple runs.

\begin{equation}
S(E_i^m, \mathbf{E}^m) \;=\;
\frac{1}{(|\mathbf{E}^m|-1)\,|E_i^m|}
\sum_{e \in E_i^m}
\sum_{\substack{E_j^m \in \mathbf{E}^m \\ j \neq i}}
\mathbf{1}_{\{e \in E_j^m\}},
\end{equation}

\noindent where $\mathbf{1}_{\{e \in E_j^m\}}$ is an indicator function that equals $1$ if 
edge $e$ appears in solution $E_j^m$, and $0$ otherwise. 
The quantity $S(E_i^m,\mathbf{E}^m)$ lies in the interval $[0,1]$, with 
$S=1$ indicating perfect stability (all edges in $E_i^m$ reappear in every other run) 
and $S=0$ indicating complete instability. Therefore, a model's stability is defined as 
\begin{eqnarray}
    \tilde{S_m} = \frac{1}{|\textbf{E}^m|}\sum_{E_i^m \in \textbf{E}^m} S(E_i^m, \mathbf{E}^m)
    \label{eq:soln_stability}
\end{eqnarray}

\noindent Convergence plots of the objective function value evolution were qualitatively evaluated as well. All median and mean scores reported in the study were considered unequal only when the Mann-Whitney U test was statistically significant. A significance level of 0.05 was used for the hypothesis tests ($H_0: \mu_{1} = \mu_{2}$) comparing the mean of each algorithm result with the best in the set. 

\section{Results}

\subsection{Edge Unconstrained Design}

\subsubsection{Quality cost performance}

\noindent Table \ref{tab:results_obj} shows the performance of each algorithm based on the average objective function values. 

\begin{table}[H]
\centering
\caption{Performance of metaheuristic techniques based on the average objective function value. Average of thirty optimisation runs. The sign for the mean comparison indicates whether the best objective value is smaller (<) or greater (>) than the best value after hypothesis testing.}
\begin{tabular}{llllllc}
\hline
\textbf{Method} & \textbf{Obj} & \textbf{Std Obj} & \textbf{Obj Min} & \textbf{Obj Max} & \textbf{n-fold} & \textbf{($\mu_1\neq \mu^*$)} \\
\hline
\textbf{Gr-SA}   &\textbf{ 47103.69 }  & 2904.39   & \textbf{41991.21 }   & \textbf{53509.91 }    & \textbf{196.50 }  & --    \\
GrA-TS  & 49461.25   & 1708.33    & 45453.33    & 54367.26     & 187.20   & Yes>  \\
SA      & 59694.55   & 22681.53   & 35790.33    & 143590.92    & 155.10   & Yes>  \\
GrA     & 66348.37   & 0.00       & 66348.37    & 66348.37     & 139.50   & Yes>  \\
GA      & 87001.42   & 32290.62   & 58249.42    & 160623.20    & 106.40   & Yes>  \\
ACO     & 115443.32  & 19321.17   & 77936.84    & 150986.43    & 80.20    & Yes>  \\
PSO     & 168127.69  & 27528.74   & 122883.92   & 227832.62    & 55.10    & Yes>  \\
TS      & 825638.52  & 843000.21  & 61920.96    & 2376160.58   & 11.20    & Yes>  \\
\hline
\end{tabular}

\label{tab:results_obj}
\end{table}

\noindent Figure \ref{fig:obj_box_plot} shows the final objective function variable of each method after all optimisation runs. The Tabu search results were excluded due to poor convergence of the objective value and significant outliers.

\begin{figure}[H]
    \centering
    \includegraphics[width=\linewidth]{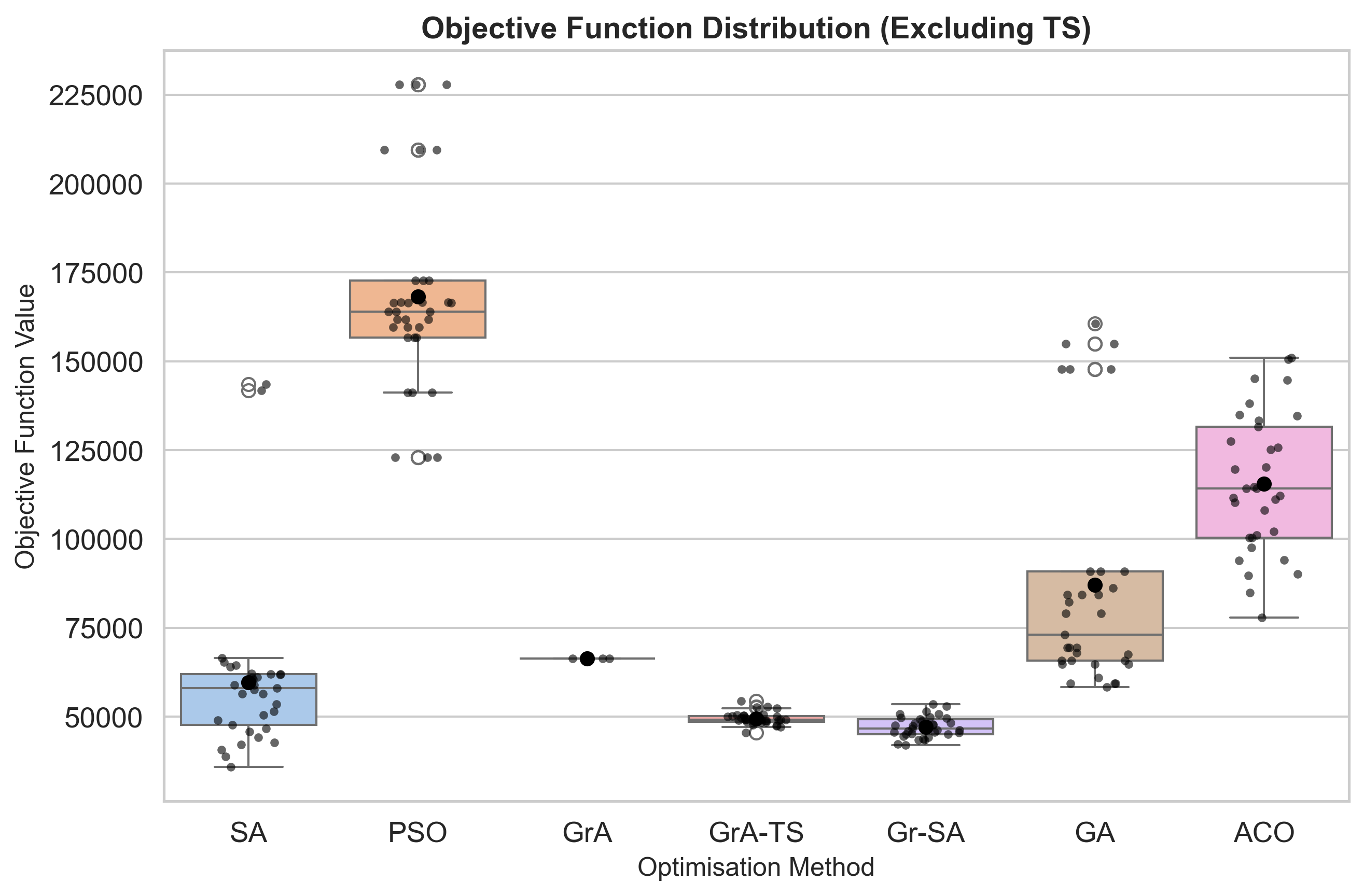}
    \caption{Objective function values per method}
    \label{fig:obj_box_plot}
\end{figure}

\begin{figure}[H]
  \centering

  % ---------- Row 1 ----------
  \begin{subfigure}{0.45\linewidth}
    \centering
    \includegraphics[width=\linewidth]{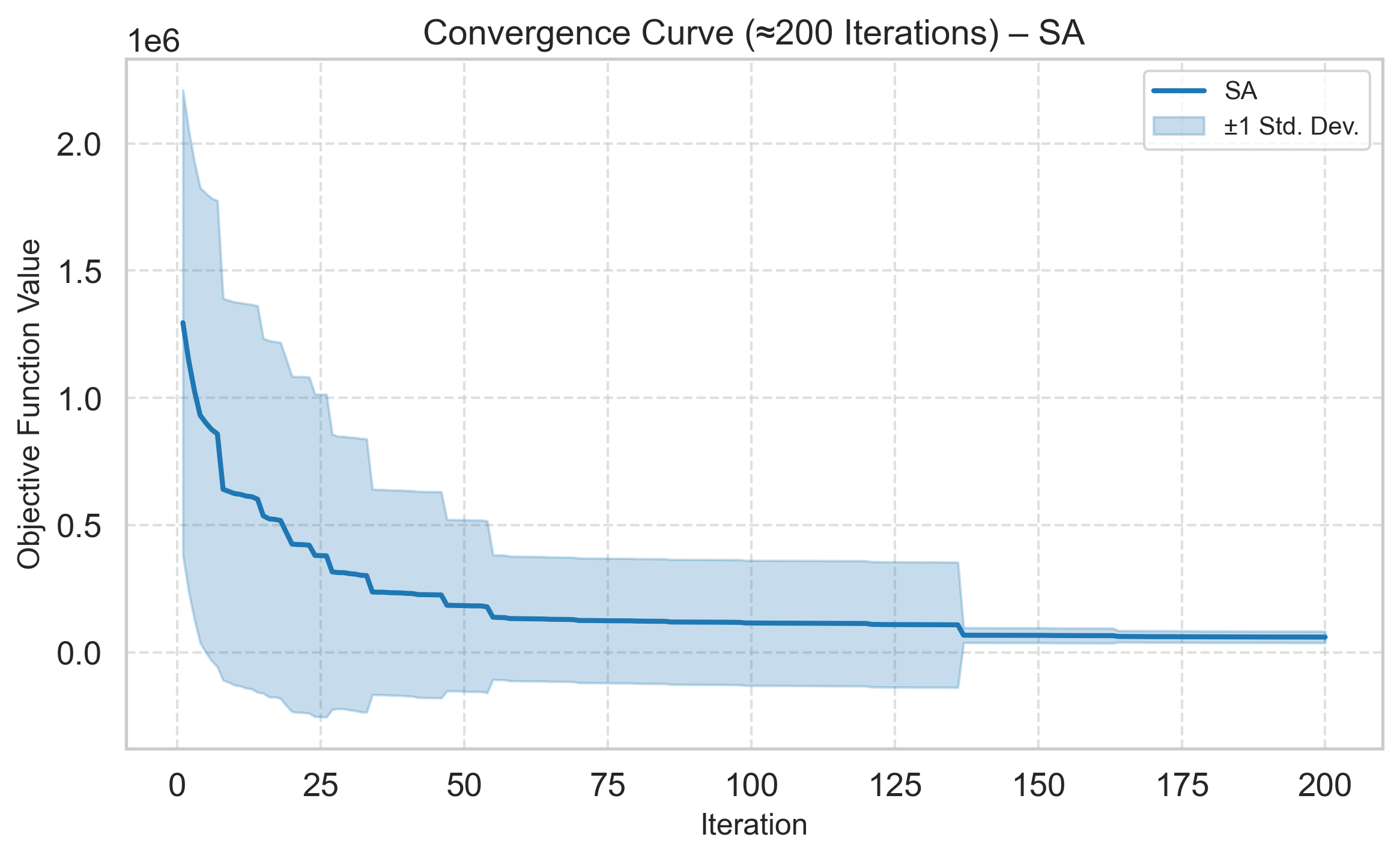}
    \caption{Simulated Annealing (SA)}
    \label{fig:conv_sa}
  \end{subfigure}
  \hfill
  \begin{subfigure}{0.45\linewidth}
    \centering
    \includegraphics[width=\linewidth]{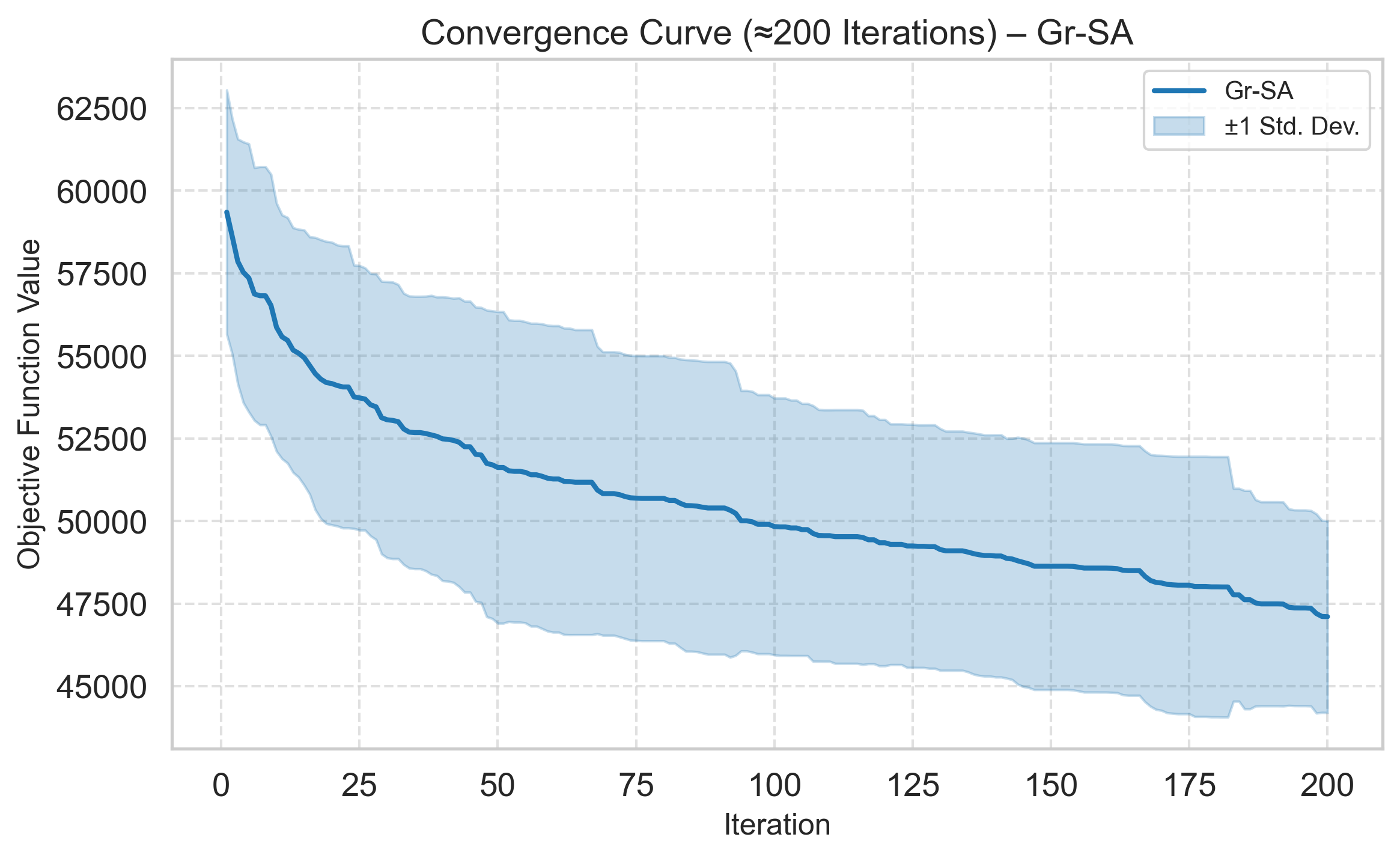}
    \caption{Greedy–Simulated Annealing (Gr-SA)}
    \label{fig:conv_grsa}
  \end{subfigure}

  % ---------- Row 2 ----------
  \vspace{0.3cm}
  \begin{subfigure}{0.45\linewidth}
    \centering
    \includegraphics[width=\linewidth]{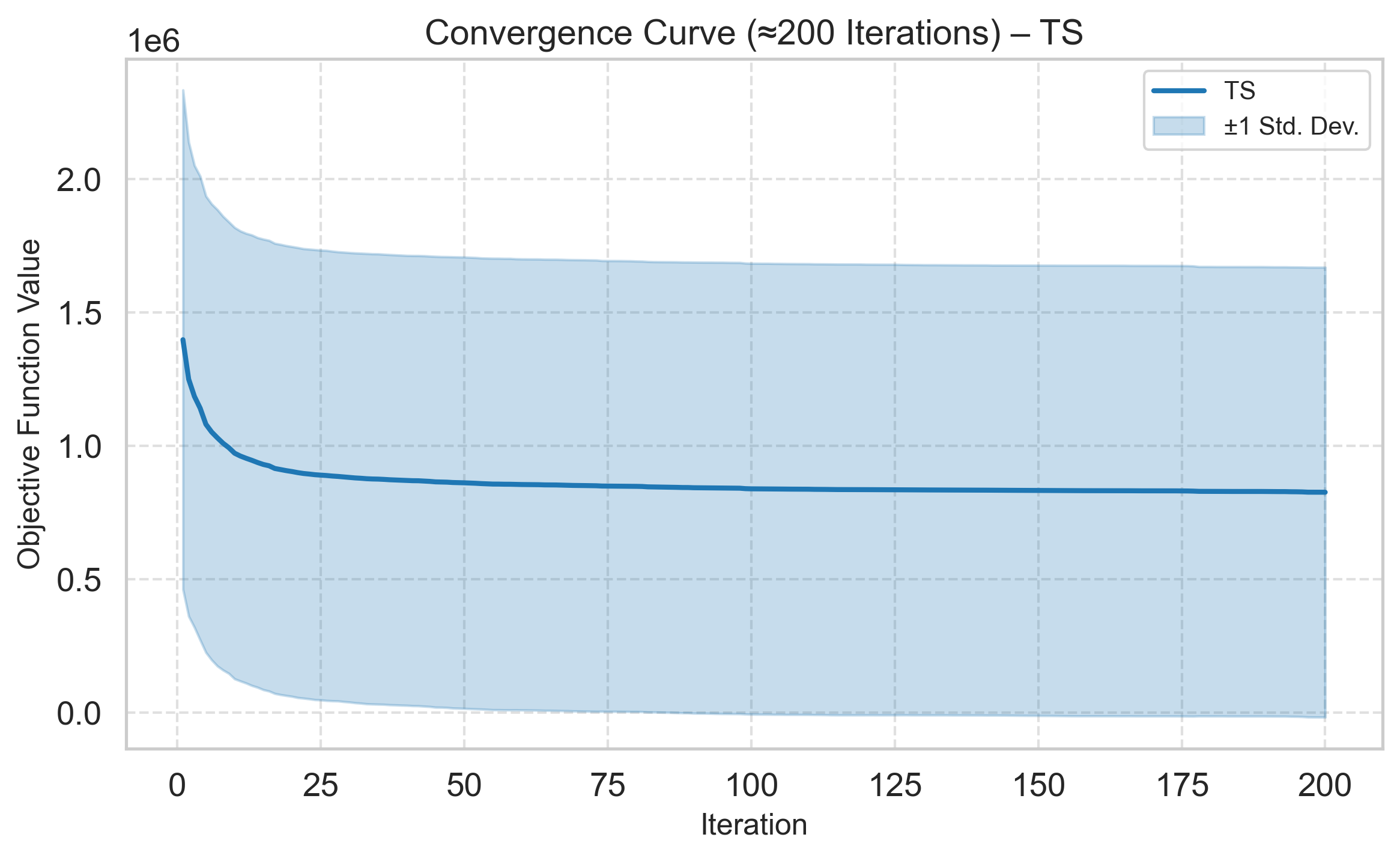}
    \caption{Tabu Search (TS)}
    \label{fig:conv_ts}
  \end{subfigure}
  \hfill
  \begin{subfigure}{0.45\linewidth}
    \centering
    \includegraphics[width=\linewidth]{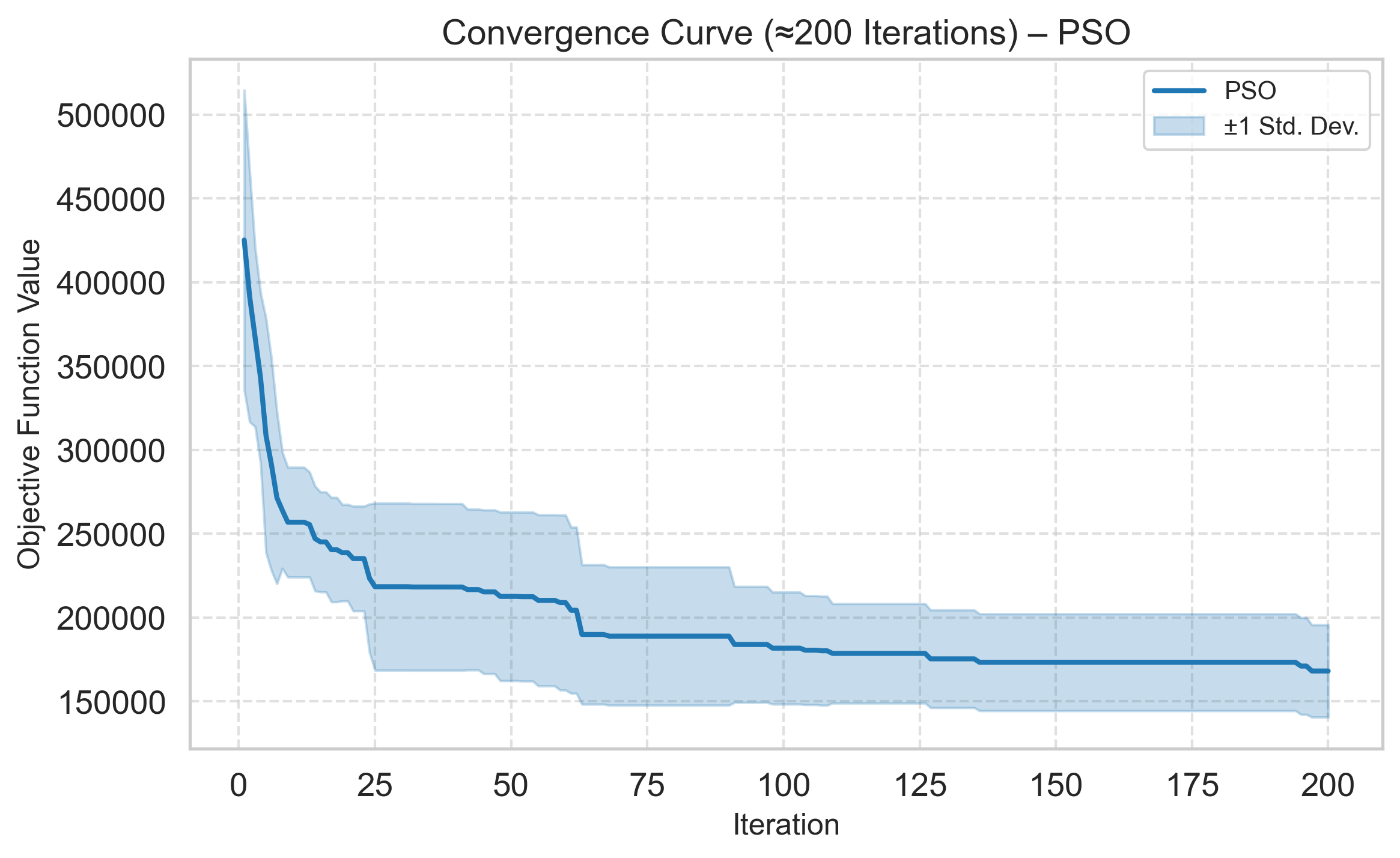}
    \caption{Particle Swarm Optimisation (PSO)}
    \label{fig:conv_pso}
  \end{subfigure}

  % ---------- Row 3 ----------
  \vspace{0.3cm}
  \begin{subfigure}{0.45\linewidth}
    \centering
    \includegraphics[width=\linewidth]{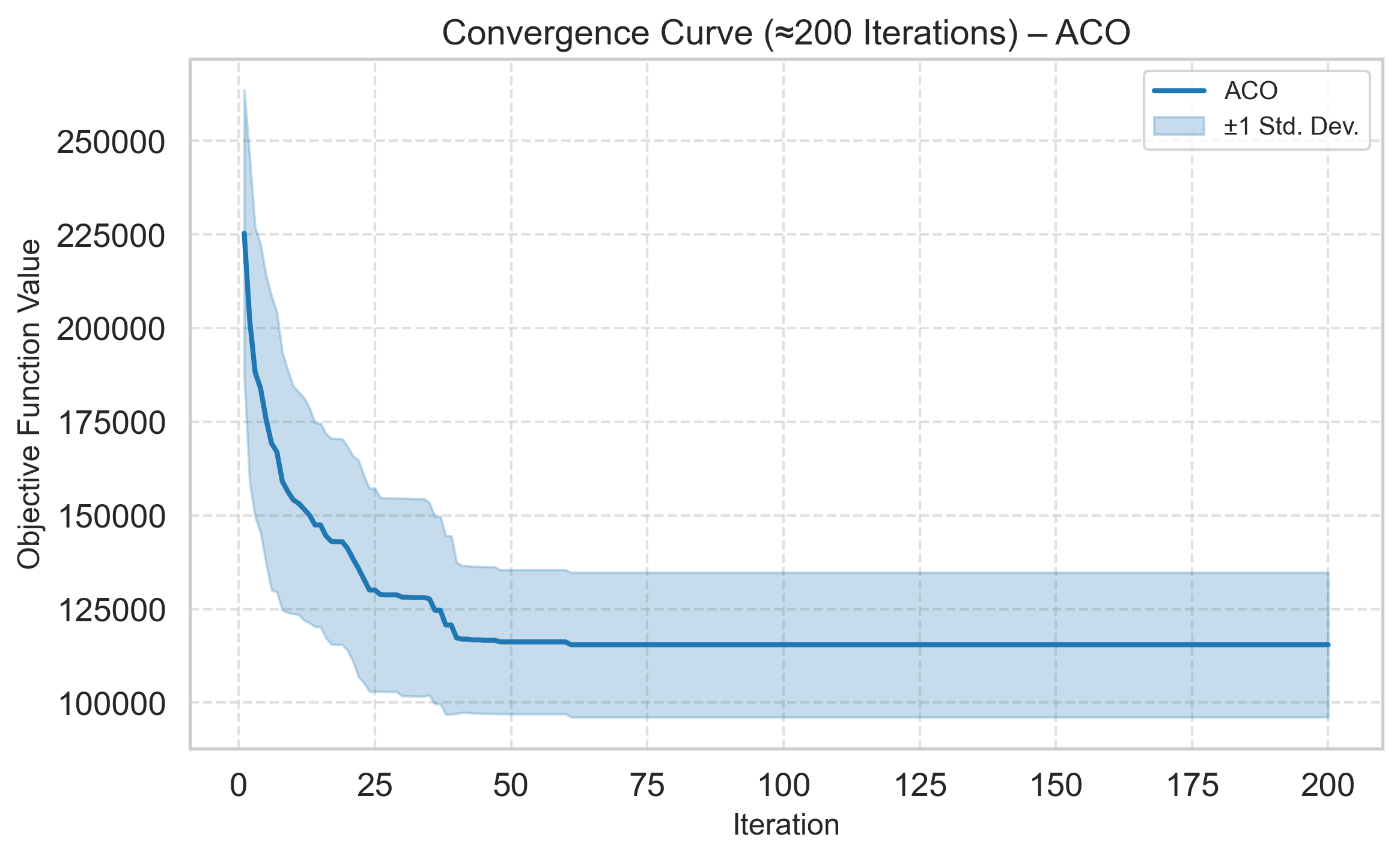}
    \caption{Ant Colony Optimisation (ACO)}
    \label{fig:conv_aco}
  \end{subfigure}
  \hfill
  \begin{subfigure}{0.45\linewidth}
    \centering
    \includegraphics[width=\linewidth]{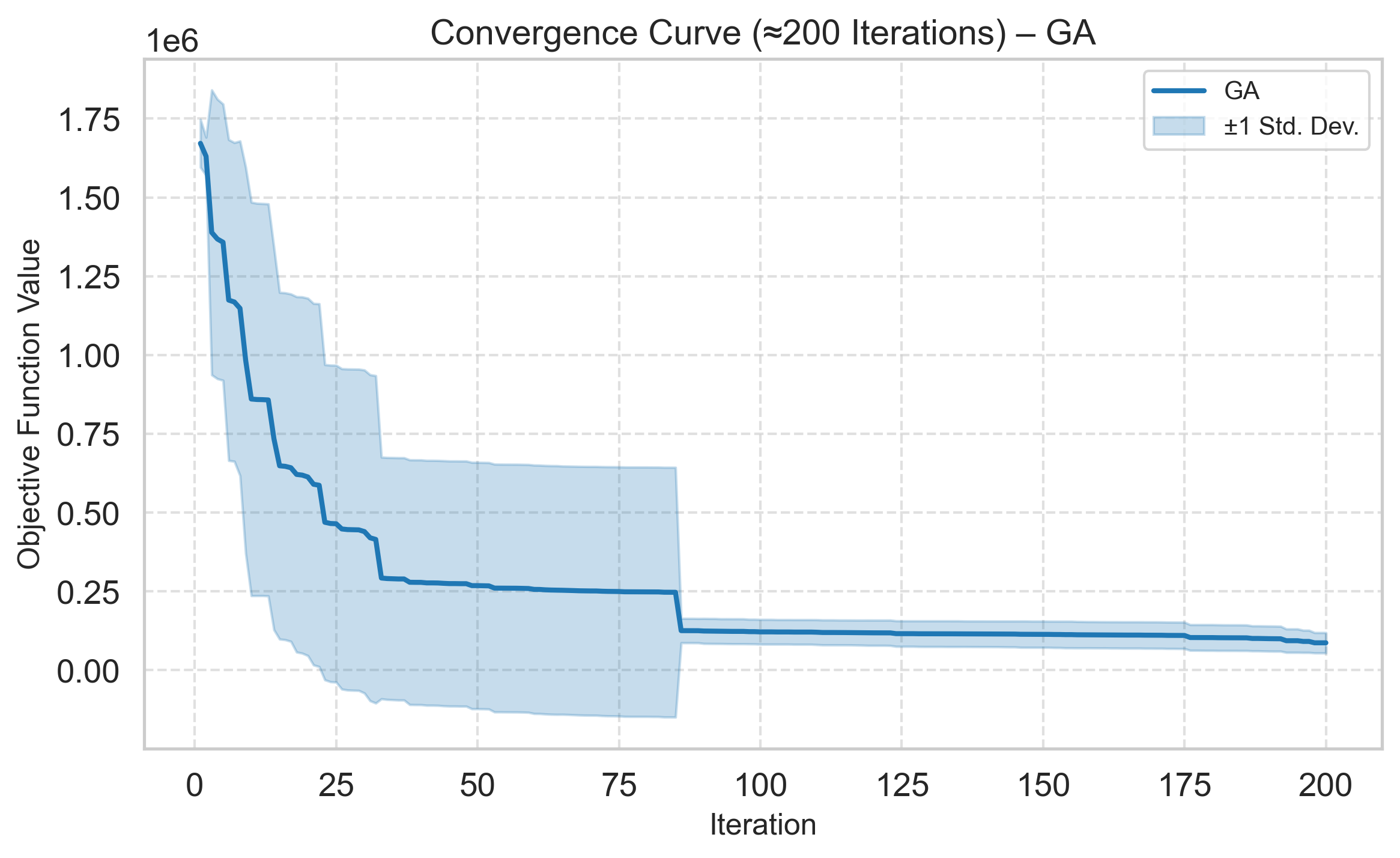}
    \caption{Genetic Algorithm (GA)}
    \label{fig:conv_ga}
  \end{subfigure}

  \caption{Convergence behaviour of metaheuristic algorithms with variance bands for the Kinshasa road-network optimisation problem. Each subfigure shows the mean objective evolution with shaded $\pm1\sigma$ regions across runs.}
  \label{fig:convergence_all}
\end{figure}

\noindent Figure \ref{fig:convergence_all} shows convergence plots of each solution algorithm as the number of iterations passes, with standard deviation variations upon multiple runs. 

\subsubsection{Solution Stability}

\begin{figure}[H]
    \centering \includegraphics[width=1.0\linewidth]{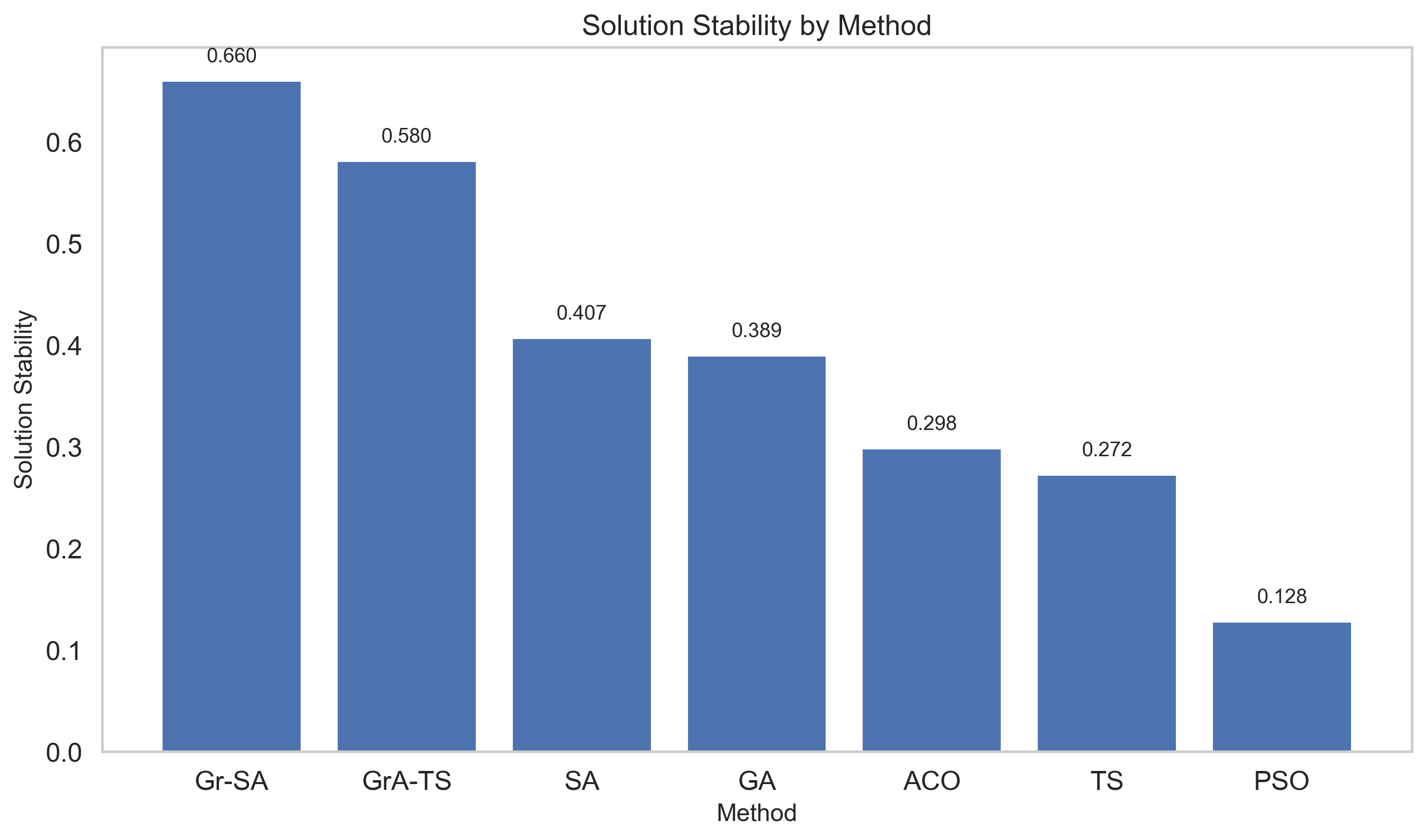}
    \caption{Solution Stability per method}
\label{fig:soln_stability}
\end{figure}

\noindent Figure \ref{fig:soln_stability} reports the solution stability of each algorithm based on the score ($\tilde{S_m} $) established in equation \ref{eq:soln_stability}.

\subsubsection{Edge Centrality betweenness}

\noindent Table \ref{tab:centr_u} reports the average edge betweenness centrality obtained from each algorithm network design benchmarked against the initial network average edge betweenness centrality.

\begin{table}[H]
\centering
\caption{Performance of metaheuristic techniques based on the average edge centrality betweenness. Average of thirty optimisation runs.}
\label{tab:results_cb}
\begin{tabular}{lrrrrrc}
\hline
\textbf{Method} & \textbf{$C_b$} & \textbf{Std $C_b$} & \textbf{$C_b$ Min} & \textbf{$C_b$ Max} & \textbf{n-fold} &\textbf{($\mu_1\neq \mu^*$)}\\
\hline
\textbf{GA }     & \textbf{0.04386}   & \textbf{0.00151  } & \textbf{0.04025 }  & 0.04659   & \textbf{3.31300 }  & --    \\
ACO     & 0.04391   & 0.00177   & 0.03990   & 0.04970   & 3.30900   & Yes>  \\
SA      & 0.04544   & 0.00251   & 0.04235   & 0.05235   & 3.19800   & Yes>  \\
PSO     & 0.05708   & 0.00373   & 0.05209   & 0.06533   & 2.54600   & Yes>  \\
Gr-SA   & 0.05849   & 0.00177   & 0.05391   & 0.06245   & 2.48400   & Yes>  \\
GrA     & 0.06159   & 0.00000   & 0.06159   & 0.06159   & 2.35900   & --    \\
GrA-TS  & 0.06225   & 0.00058   & 0.06135   & 0.06341   & 2.33400   & Yes>  \\
TS      & 0.06583   & 0.00660   & 0.05093   & 0.08452   & 2.20700   & Yes>  \\
\hline
\end{tabular}
\label{tab:centr_u}
\end{table}

\subsubsection{Execution time}

\noindent Table \ref{tab:results_time} shows the performance of each algorithm based on the average execution time, accompanied by a similar box plot (Figure \ref{fig:computation_time}).

\begin{table}[H]
\centering
\caption{Performance of metaheuristic techniques based on the average computation time. Average of thirty optimisation runs.}
\label{tab:results_time}
\begin{tabular}{lrrrrc}
\hline
\textbf{Method} & \textbf{t\_eps} & \textbf{Std} & \textbf{t\_min} & \textbf{t\_max} & \textbf{($\mu_1\neq \mu^*$)}\\
\hline
\textbf{GrA}     & \textbf{27.34  } & \textbf{0.00 }    & \textbf{27.34 }    & \textbf{27.34}     & --    \\
\textbf{GrA-TS } & \textbf{118.51 } & \textbf{104.35  } & \textbf{113.39 }   & \textbf{538.45}    & --    \\
Gr-SA   & 118.71  & 116.76   & 98.84     & 577.23    & No    \\
TS      & 119.18  & 105.22   & 108.32    & 531.62    & No    \\
SA      & 121.94  & 118.48   & 75.86     & 583.00    & Yes>  \\
GA      & 134.55  & 113.17   & 45.26     & 547.84    & Yes>  \\
PSO     & 135.02  & 114.27   & 43.59     & 553.17    & Yes>  \\
ACO     & 141.44  & 116.79   & 31.48     & 557.46    & Yes>  \\
\hline
\end{tabular}
\label{tab:compute_time_u}
\end{table}

\begin{figure}[H]
    \centering
    \includegraphics[width=\linewidth]{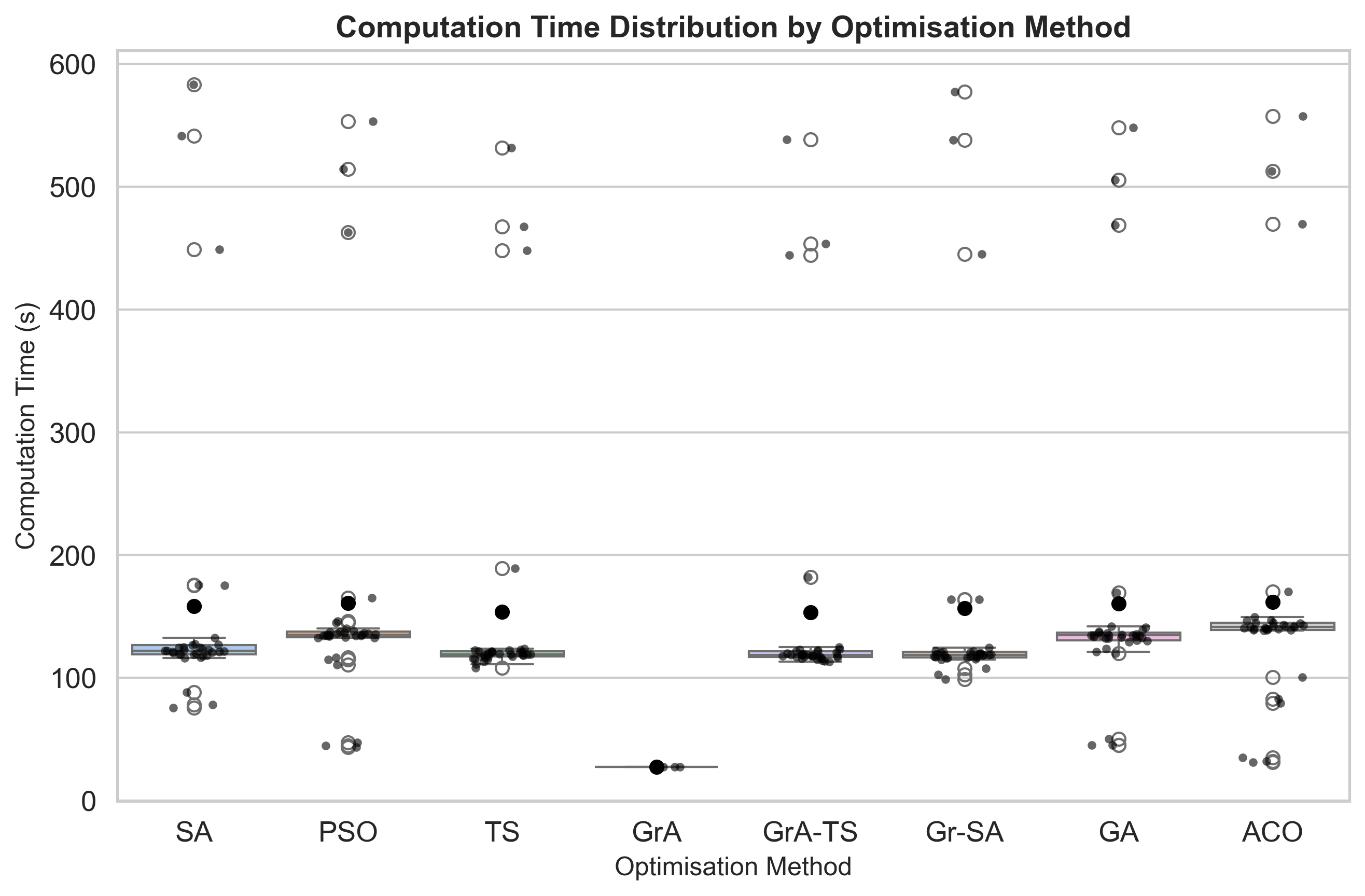}
    \caption{Computation time of all algorithms across optimisation runs}
    \label{fig:computation_time_u}
\end{figure}

\subsection{Edge Constrained Design}

\subsubsection{Quality cost performance}
\begin{figure}[H]
    \centering
    \begin{subfigure}{0.9\linewidth}
        \centering \includegraphics[width=\linewidth]{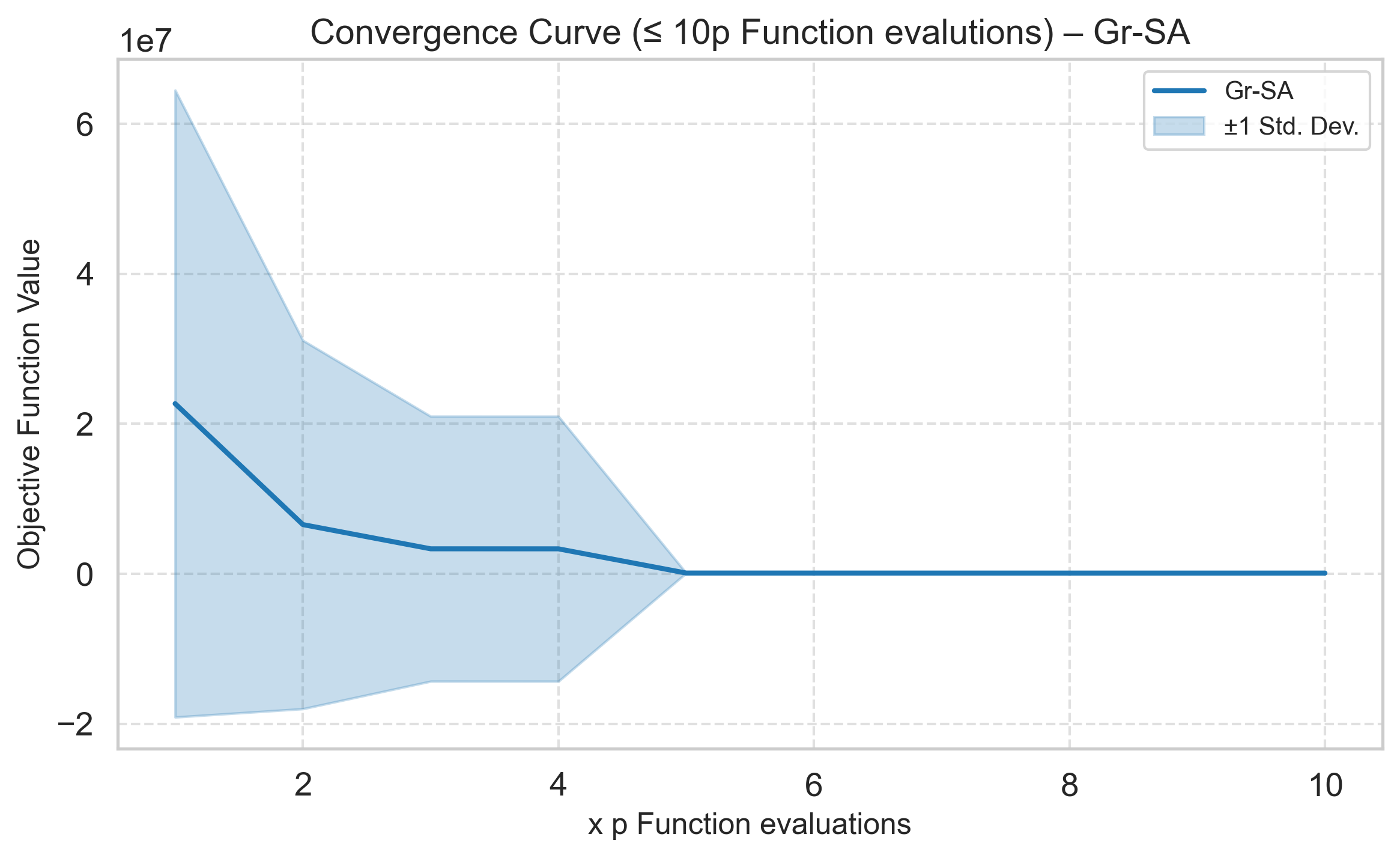}
        \caption{Convergence plot of Gr-SA}
        \label{fig:constr_gr_sa}
    \end{subfigure}

    \vspace{0.4cm}

    \begin{subfigure}{0.9\linewidth}
        \centering
        \includegraphics[width=\linewidth]{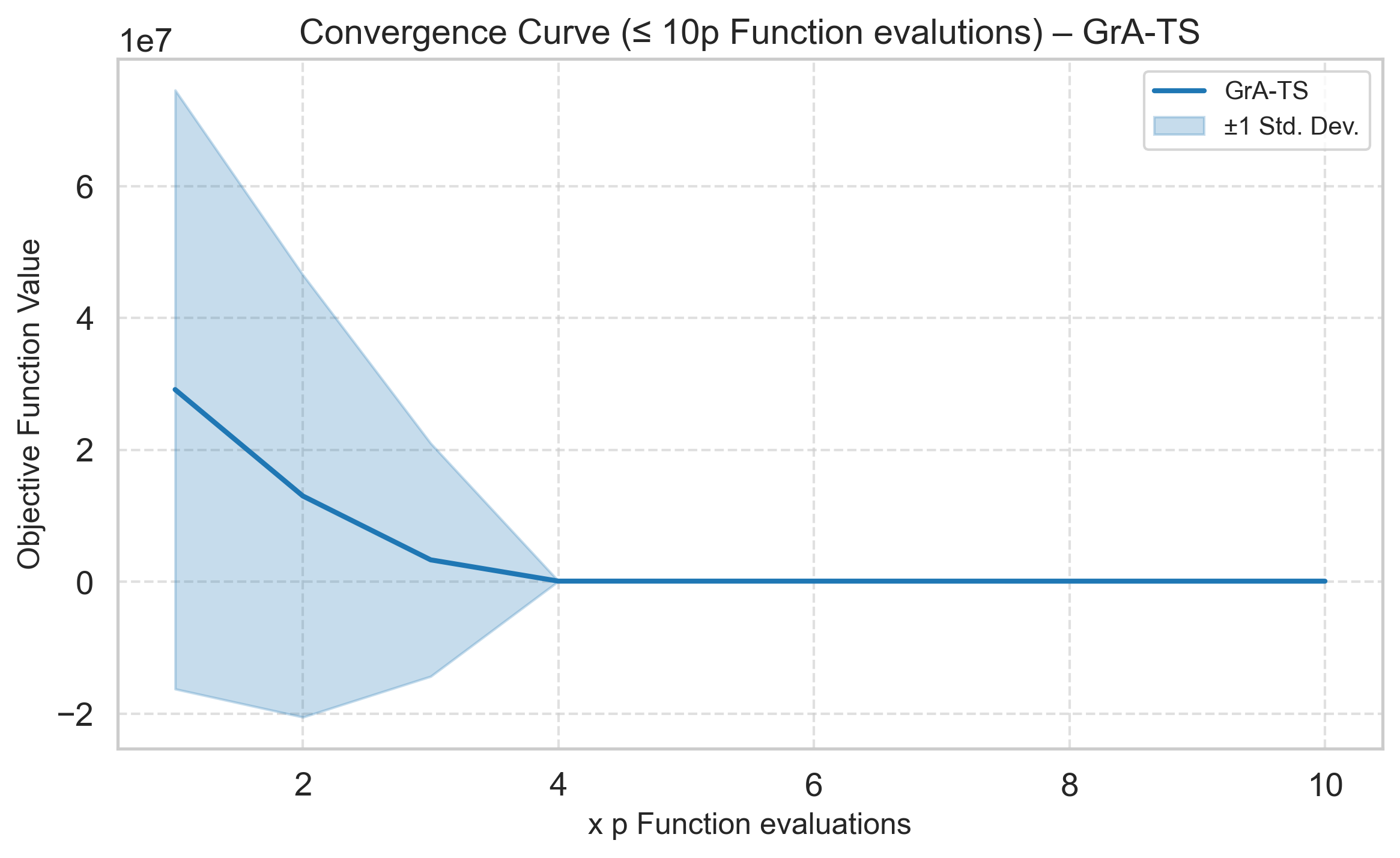}
        \caption{Convergence plot of GrA-TS}
        \label{fig:constr_gr_ts}
    \end{subfigure}

    \caption{Convergence comparison of hybrid methods Gr-SA and GrA-TS for constrained design}
    \label{fig:constr_vertical_comparison_c}
\end{figure}

\noindent Table \ref{tab:results_obj_c} and  Figure \ref{fig:obj_box_plot} report the objective values for constrained design algorithms, including results from objective values convergence plots (See Figure \ref{fig:constr_vertical_comparison_c}). Only algorithms that yielded superior results in the constrained design were used in this phase.

\begin{table}[H]
\centering
\caption{Performance of metaheuristic techniques based on the average objective function value. Average of thirty optimisation runs.}
\begin{tabular}{llllllc}
\hline
\textbf{Method} & \textbf{Obj} & \textbf{Std Obj} & \textbf{Obj Min} & \textbf{Obj Max} &\textbf{ n-fold} & ($\mu_1\neq \mu^*$)\\
\hline
\textbf{Gr-SA  } & \textbf{87841.28 }  & 8799.71    & \textbf{66376.78 }   & \textbf{98894.43 }    & \textbf{105.40 }  & --    \\
GrA-TS  & 96318.48   & \textbf{6616.23 }   & 84714.53    & 119368.92    & 96.10    & Yes>  \\
GrA     & 102039.03  & 0.00       & 102039.03   & 102039.03    & 90.70    & No    \\
\hline
\end{tabular}

\label{tab:results_obj_c}
\end{table}

\begin{figure}[h]
    \centering
    \includegraphics[width=\linewidth]{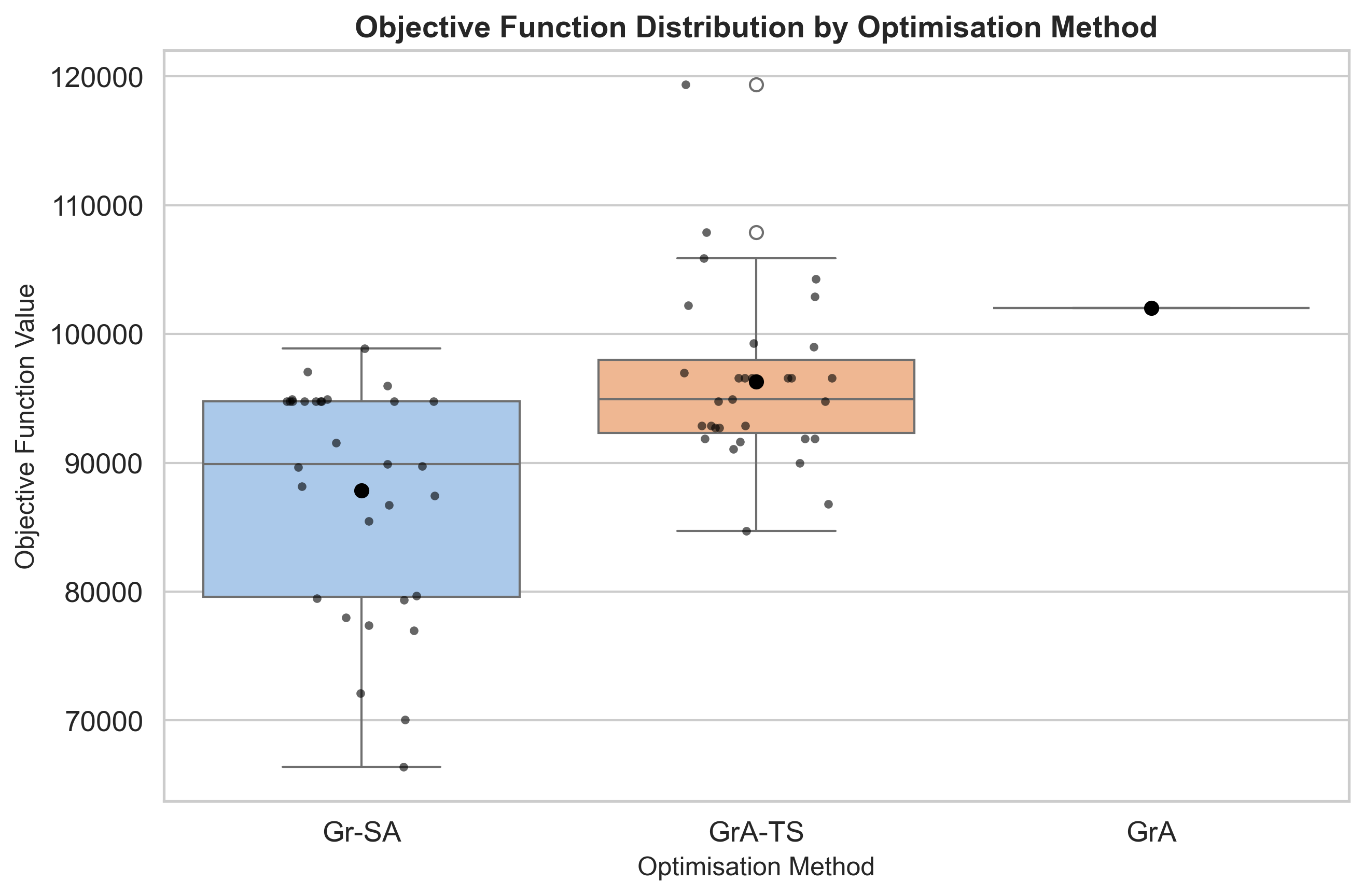}
    \caption{Objective function values per method}
    \label{fig:obj_box_plot}
\end{figure}

\subsubsection{Solution Stability and Edge Centrality Betweenness}

\noindent Tables \ref{tab:soln_stab_c} and \ref{tab:edge_between_c} report, respectively, the solution stability of the constrained design methods and the resulting average edge betweenness centralities. 

\begin{table}[H]
\caption{Solution stability per method}
\label{tab:constrined_sol_stability}
\centering
\begin{tabular}{lr}
\hline
\textbf{method} & \textbf{solution stability}\\
\hline

GrA-TS & 0.75355 \\
Gr-SA  & 0.74973 \\
\hline
\end{tabular}
\label{tab:soln_stab_c}
\end{table}

\begin{table}[H]
\centering
\caption{Performance of metaheuristic techniques based on the average edge betweenness centrality. Average of thirty optimisation runs.}
\label{tab:results_time}
\begin{tabular}{lrrrrrc}
\hline
\textbf{Method} & \textbf{$C_b$} & \textbf{Std $C_b$} & \textbf{$C_b$ Min} & \textbf{$C_b$ Max} & \textbf{n-fold} & $(\mu_1\neq \mu_2)$ \\
\hline
\textbf{GrA }    & \textbf{0.05342}   & \textbf{0.00000 }  & 0.05342   & 0.05342   & \textbf{2.72000}   & --    \\
Gr-SA   & 0.05401   & 0.00174   & \textbf{0.04990}   & 0.05643   & 2.69000   & No    \\
GrA-TS  & 0.05543   & 0.00117   & 0.05342   & 0.05840   & 2.62000   & No    \\
\hline
\end{tabular}
\label{tab:edge_between_c}
\end{table}

\subsubsection{Execution time}

\begin{figure}[H]
    \centering
    \includegraphics[width=\linewidth]{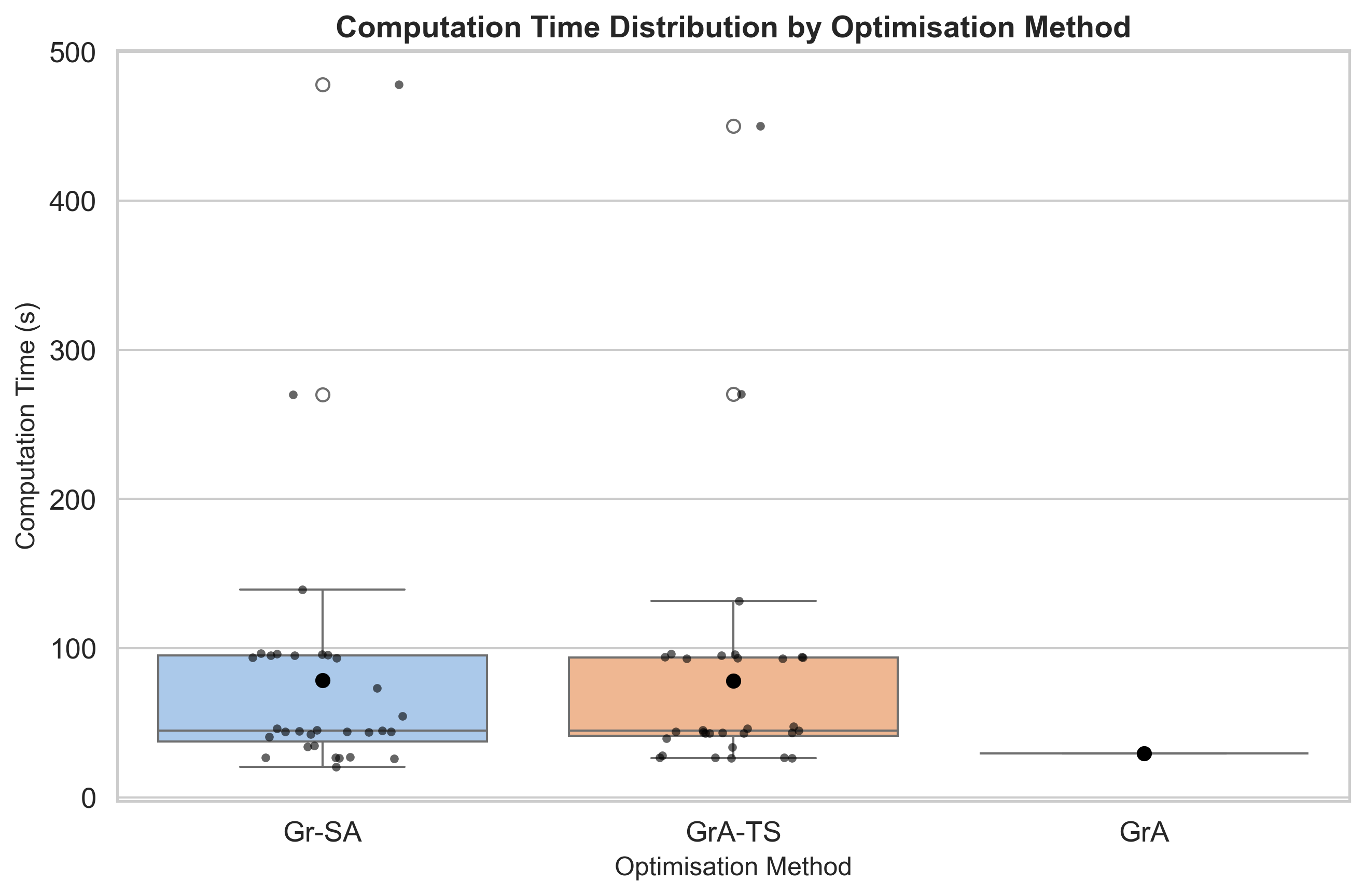}
    \caption{Computation time of all algorithms across optimisation runs}
    \label{fig:computation_time}
\end{figure}

\noindent Table \ref{tab:results_time} shows the performance of each algorithm based on the average execution time, accompanied by a similar box plot (Figure \ref{fig:computation_time}).

\begin{table}[H]
\centering
\caption{Performance of metaheuristic techniques based on the average computation time. Average of thirty optimisation runs.}
\label{tab:results_time}
\begin{tabular}{lrrrrc}
\hline
\textbf{Method} & \textbf{t\_eps} & \textbf{Std} & \textbf{t\_min} & \textbf{t\_max} & \textbf{($\mu_1\neq \mu^*$)} \\
\hline
\textbf{GrA }    & \textbf{29.51 }  & \textbf{0.00 }   &\textbf{ 29.51 }   & \textbf{29.51  }   & -    \\
\textbf{GrA-TS}  & \textbf{44.58  } & \textbf{82.82 }  & \textbf{26.33}    & \textbf{450.17 }   & -   \\
Gr-SA   & 44.74   & 87.15   & 20.25    & 477.89    & No    \\
\hline
\end{tabular}
\end{table}

\section{Discussion and Policy Recommendations}
\subsection{Hybrid Greedy-Local Exploration yields superior network design performance}
\noindent The research design philosophy was centred on a priority-based near-deterministic construction methodology inherent to greedy search supplemented with local exploration; Greedy search yielded an initial, efficient, priority-informed good approximation, combined with the Simulated annealing and Tabu perturbation algorithm. This design philosophy yielded superior computational results, with the hybrid Greedy-Simulated annealing and Greedy-Tabu search achieving the highest reductions in travel time (See Tables \ref{tab:results_obj}, \ref{tab:results_obj_c} and Figures \ref{fig:obj_box_plot}, \ref{fig:constr_vertical_comparison_c} i.e., unconstrained: 47104/196.50 and constrained: 49461/187.20 - 87841/105 and 96318/96), thereby encouraging our design approach. Unconstrained design yielded the largest reduction in travel time compared to the constrained design; however, it entailed greater construction demands. Genetic algorithms, Ant colony, and Simulated annealing yielded the highest average edge betweenness centrality reduction relative to the original network (See Figure; i.e., unconstrained: 3.13, 3.31, 3.20 against 2.48 and 2.33 for Gr-SA, Gr-TS). While network centrality triples with GA, ACO, and SA, their travel times are not significantly reduced, leaving the two-and-a-half-fold reduction in edge betweenness centrality by Gr-SA and Gr-TS viable. Congestion minimisation based on demand remains the natural, realistic target. 

\par The magnitude of the travel-time improvements—ranging from approximately 11-fold to nearly 200-fold—reflects the extreme level of congestion in Kinshasa’s baseline road network and the nonlinear nature of the BPR function, where travel time increases sharply as flow approaches capacity. These values are also influenced by the estimation and normalisation of OD demand data in the absence of publicly available traffic measurements. Despite variations in absolute magnitudes, the relative ranking of algorithms remains consistent, strengthening confidence in the comparative conclusions.

\subsection{On the Stability of the Edge Solution Set and Computational Considerations}
\noindent Using the solution stability score presented in equation \ref{eq:soln_stability}, the greedy-simulated annealing and greedy-tabu search yielded the highest stability score in both the unconstrained and constrained design (See Figure \ref{fig:soln_stability} and Table \ref{tab:soln_stab_c}:  0.660 - 0.580 and 0.75355-0.74973), strengthening the rationale of our design philosophy. Improved objective function values accompanied the high solution stability compared against other independent solvers (GA, PSO, ACO, SA). These results were obtained with minimal algorithmic re-adaptation, which could otherwise require significant modification of conventional metaheuristics (GA, PSO, ACO). 

\par Convergence plots in Figure \ref{fig:convergence_all} show stable convergence of each algorithm except Tabu search, which yielded unstable convergence when used without a seed. Investigations on other, more appropriate variants of the algorithm can be conducted for the TNDP. Nevertheless, when seeded with a Greedy solution, the algorithm exhibited good convergence and superior performance (See Table \ref{tab:results_obj} and Figure \ref{fig:constr_gr_ts}). In the unconstrained design, GA exhibited early convergence, whereas Gr-SA suggested further solution improvement with an increased maximum iteration count.
The hybrid Gr-SA and Gr-TS converged quickly during constrained design after approximately 80 to 100 function evaluations (i.e., $p=20$) upon obtaining the greedy solution seed. On the computation time, all algorithms besides the Greedy algorithm yielded comparable execution time (See Figure \ref{fig:computation_time} and Table \ref{tab:compute_time_u}), given that they were evaluated with the same number of function evaluations and minimal algorithmic overhead.

\subsection{Road construction recommendations for the city of Kinshasa}
\begin{figure}[H]
    \centering
\includegraphics[width=\linewidth,height=8cm]{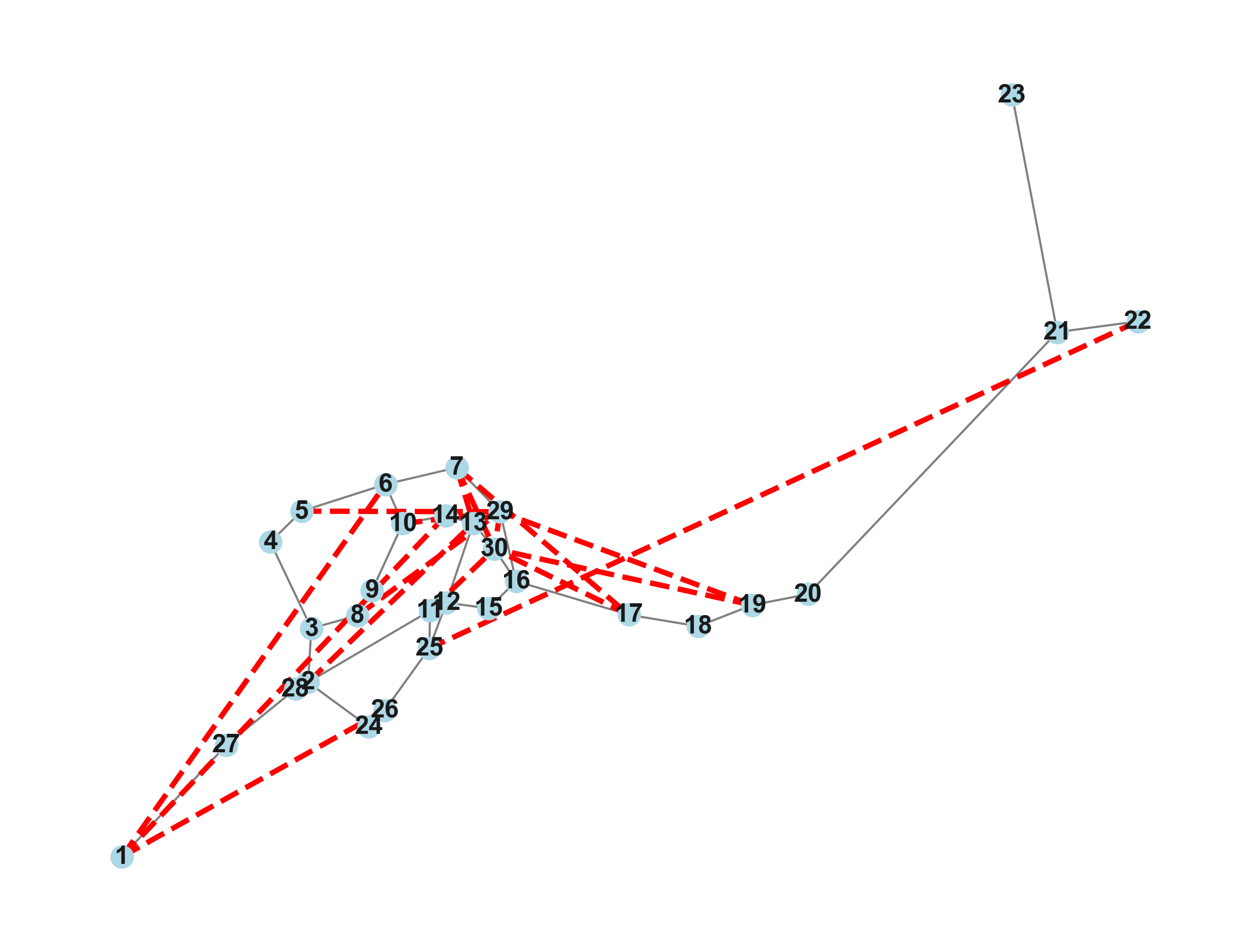}
    \caption{Constrain-Free Augmented Network Design}
    \label{fig:free_design_recommendations}
\end{figure}

\begin{figure}[H]
    \centering
\includegraphics[width=\linewidth,height=9cm]{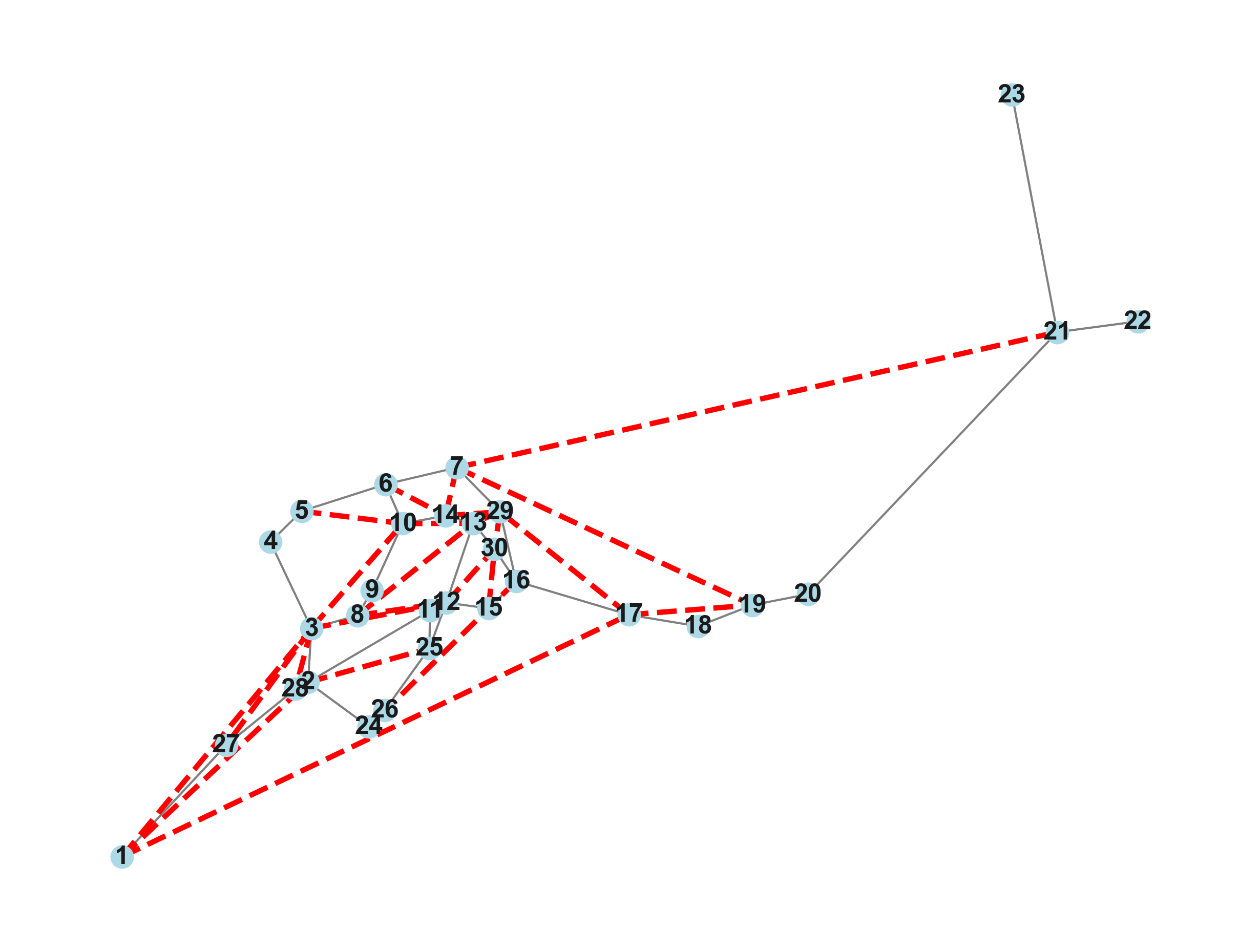}
    \caption{Constrained Augmented Network Design}
    \label{fig:constr_design_recommendations}
\end{figure}

\noindent The computational experiment comprised two design strategies: an unconstrained design that typically proposes the main junctions regardless of geographic road intersections (Figure \ref{fig:free_design_recommendations}) and a constrained design that connects road central nodes without permitting intersections (Figure \ref{fig:constr_design_recommendations}). The former design strategy provides a skeletal construction philosophy that advises on which areas need to be joined, and the second design strategy provides a practical, constrained construction plan for the city of Kinshasa.\clearpage

\begin{table}[H]
\centering
\caption{Road Recommendations Results}
\label{tab:results_recommendation}
\scalebox{0.9}{
\begin{tabular}{|c|l|c|l|c|}
\hline
\textbf{Node 1} & \textbf{Description} & \textbf{Node 2} & \textbf{Description} & \textbf{km ($\approx$)} \\
\hline
1  & Route de Matadi (border Kinshasa) & 3  & UPN                       & 17.69 \\ \hline
17 & Masina                             & 29 & Limete Poids Lourd        & 9.05  \\ \hline
16 & Échangeur 1                        & 24 & Arrêt Gare                & 11.89 \\ \hline
1  & Route de Matadi (border Kinshasa) & 17 & Masina                    & 27.84 \\ \hline
1  & Route de Matadi (border Kinshasa) & 2  & Triangle Matadi Kibala    & 14.47 \\ \hline
7  & Gare Centrale                      & 19 & Aéroport Ndjili           & 16.09 \\ \hline
12 & Rond Point Ngaba                   & 30 & Limete Résidentiel        & 4.24  \\ \hline
29 & Limete Poids Lourd                 & 30 & Limete Résidentiel        & 2.55  \\ \hline
10 & Pierre Mulele                      & 13 & Av. de l'Université       & 3.12  \\ \hline
5  & Mont Ngaliema                      & 10 & Pierre Mulele             & 4.55  \\ \hline
15 & Lemba                              & 30 & Limete Résidentiel        & 4.07  \\ \hline
2  & Triangle Matadi Kibala             & 25 & UNIKIN                    & 5.85  \\ \hline
17 & Masina                             & 19 & Aéroport Ndjili           & 5.50  \\ \hline
13 & Av. de l'Université                & 29 & Limete Poids Lourd        & 1.42  \\ \hline
3  & UPN                                & 27 & Benseke                   & 8.82  \\ \hline
14 & Bd Triomphal                       & 29 & Limete Poids Lourd        & 2.43  \\ \hline
7  & Gare Centrale                      & 14 & Bd Triomphal              & 3.33  \\ \hline
8  & Selembao (Auto Stop)               & 13 & Av. de l'Université       & 8.11  \\ \hline
3  & UPN                                & 10 & Pierre Mulele             & 8.22  \\ \hline
8  & Selembao (Auto Stop)               & 12 & Rond Point Ngaba          & 4.02  \\ \hline
3  & UPN                                & 28 & Wenze Matadi Kibala      & 4.20  \\ \hline
7  & Gare Centrale                      & 21 & RP Nsele                  & 28.12 \\ \hline
3  & UPN                                & 12 & Rond Point Ngaba          & 6.22  \\ \hline
6  & Bd du 30 Juin                      & 14 & Bd Triomphal              & 3.41  \\ \hline
\end{tabular}
}
\end{table}
\par The current design in Figure \ref{fig:constr_design_recommendations} advises node augmentation in three main centre areas: the entry regions to Kinshasa, Bandundu and Kongo-Central, and the inner cities, as presented in Table \ref{tab:results_recommendation}. The traffic flow from the two main road entries to Kinshasa should be decongested on their way to the main destinations: four main road constructions from the Matadi road entry (Node 1), one junction joining the Matadi entry to Masina, another junction bypassing traffic to the Matadi Kibala market centre, an additional road directly joining the UPN market centre, and two additional junctions joining Benseke to UPN directly and Matadi Kibala to UPN directly.

\par The second large traffic zone involves the Bandundy entry point to Kinshasa: a direct road construction connecting the Nsele crossroads to the city of Gombe, a direct road construction connecting the Gombe endpoint to Masina, and another that heads directly to the airport. A direct junction linking the Limete Poids Lourd area to the Masina centre and a bypass road linking the Masina centre to the Airport.

\par Finally, additional roads are suggested in the Ngaliema, Gombe-Lingwala, Ngaba-Lemba and Kimwenza centres, such as the construction or relief of a road joining the cité verte area to UNIKIN, the construction of a road joining the end of Kimwenza to Lemba, the construction of a direct road linking UPN to the Ngaba crossroad, the construction of a road linking Selembao directly to the Sendwe-Triomphal Boulevard, as displayed in Table \ref{tab:results_recommendation}. Note that these estimates do not account for physical and environmental constraints that practical construction must consider.

\par The reported significant reduction in travel time and superior solution stability of Greedy-Simulated and Greedy-Tabu search in  Tables \ref{tab:results_obj}, \ref{tab:results_obj_c} and Figures \ref{fig:obj_box_plot}, \ref{fig:constr_vertical_comparison_c} support the viability of the current design. The road recommendations proposed in Table \ref{tab:results_recommendation} can serve as informative guidelines for policymakers in the city of Kinshasa. Future research should investigate the applicability of exact methods to large network design \cite{rey2020computational}, thereby yielding deterministically rigid stable design solutions if achievable.

\section{Conclusion}
\noindent The current study proposed an optimisation-based network augmentation scheme to reduce traffic congestion in the city of Kinshasa. The Kinshasa traffic problem has been modelled as a standard discrete network problem using estimated city-origin demand data and optimised using greedy local solvers, namely Greedy-Simulated Annealing and Greedy-Tabu Search. This yielded the highest reduction in congestion time, the most stable solution profile relevant for infrastructure policymaking, and a more than two-fold improvement in network edge betweenness centrality. The design recommendation suggests three central regions of relief road additions: new roads between the Bandundu entry, passing through the airport to the Kinshasa Gombe-Limete corporate and industrial centre, roads connecting the Kongo Central entry point respectively, to the Masina area, Limete industrial region,  UPN-Matadi Kibala transit and market centres, and additional inner cities interconnection roads in the Selembao, Ngaba, Gombe, Ngaliema and Lemba communes. Future research will investigate the use of exact methods to solve the discrete TNDP, to obtain guaranteed, stable, and repeatable design solutions. Additional environmental and sociological constraints will be investigated in upcoming studies.

\section*{Acknowledgements}

\noindent This research was supported by the Tshwane University of Technology Postdoctoral Research Program at the Soshanguve South Campus, Department of Computer Systems Engineering.

\section*{Data and Code Availability}

\noindent All data and code used in this study are available in the under-development KANISA library on GitHub \cite{matanga2025kanisa_transportation}.

\bibliography{mybibfile}
% \clearpage
\section*{Appendix A -  Origin Demand - Data Collection}

%#Page 59

\begin{figure}[H]
    \centering
    \begin{subfigure}[b]{0.32\linewidth}
        \centering
        \includegraphics[width=\linewidth]{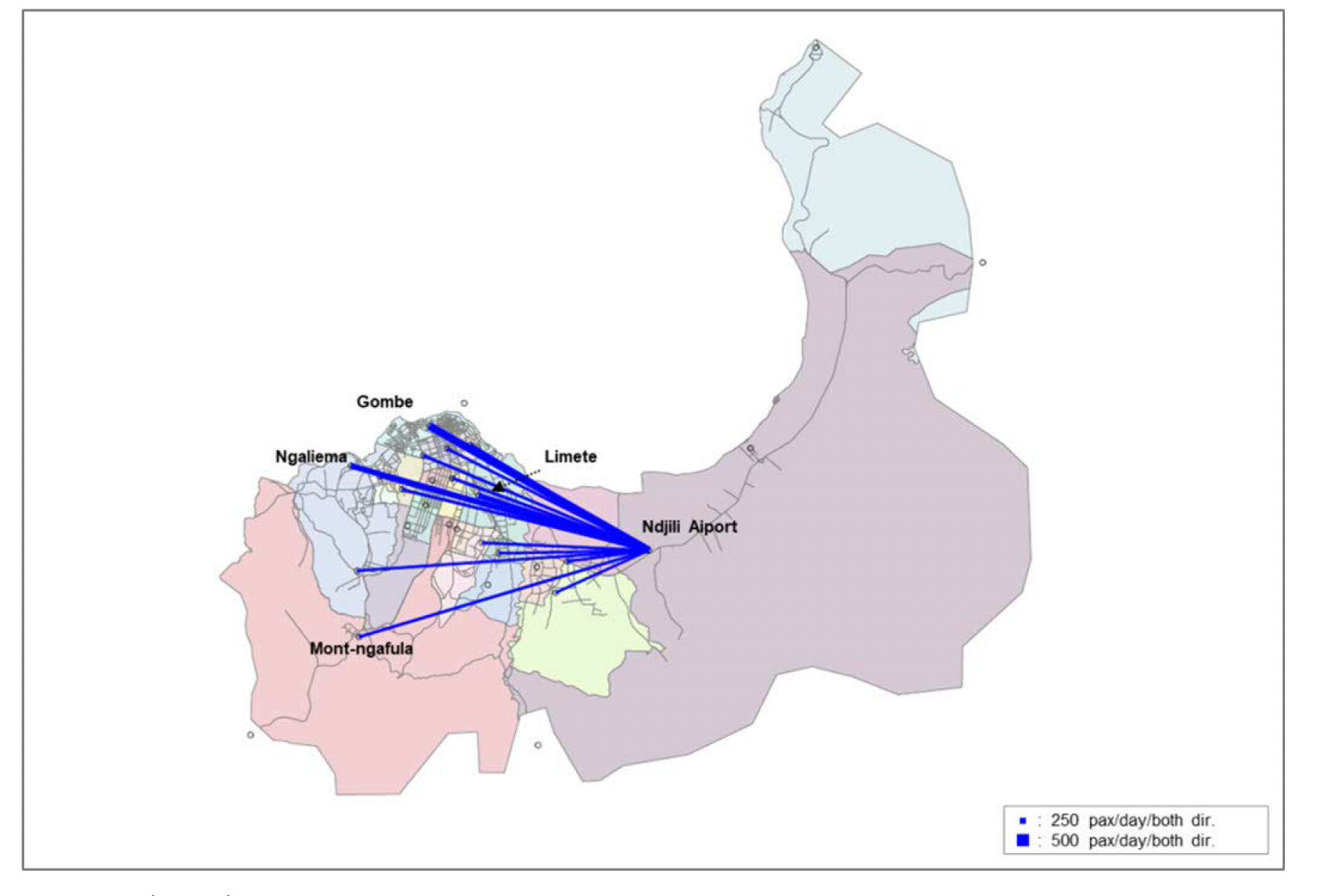}
        \caption{Traffic Main Origin Destination Pairs (Set 1)}
        \label{fig:traffic_data1}
    \end{subfigure}
    \hfill
    \begin{subfigure}[b]{0.32\linewidth}
        \centering
        \includegraphics[width=\linewidth]{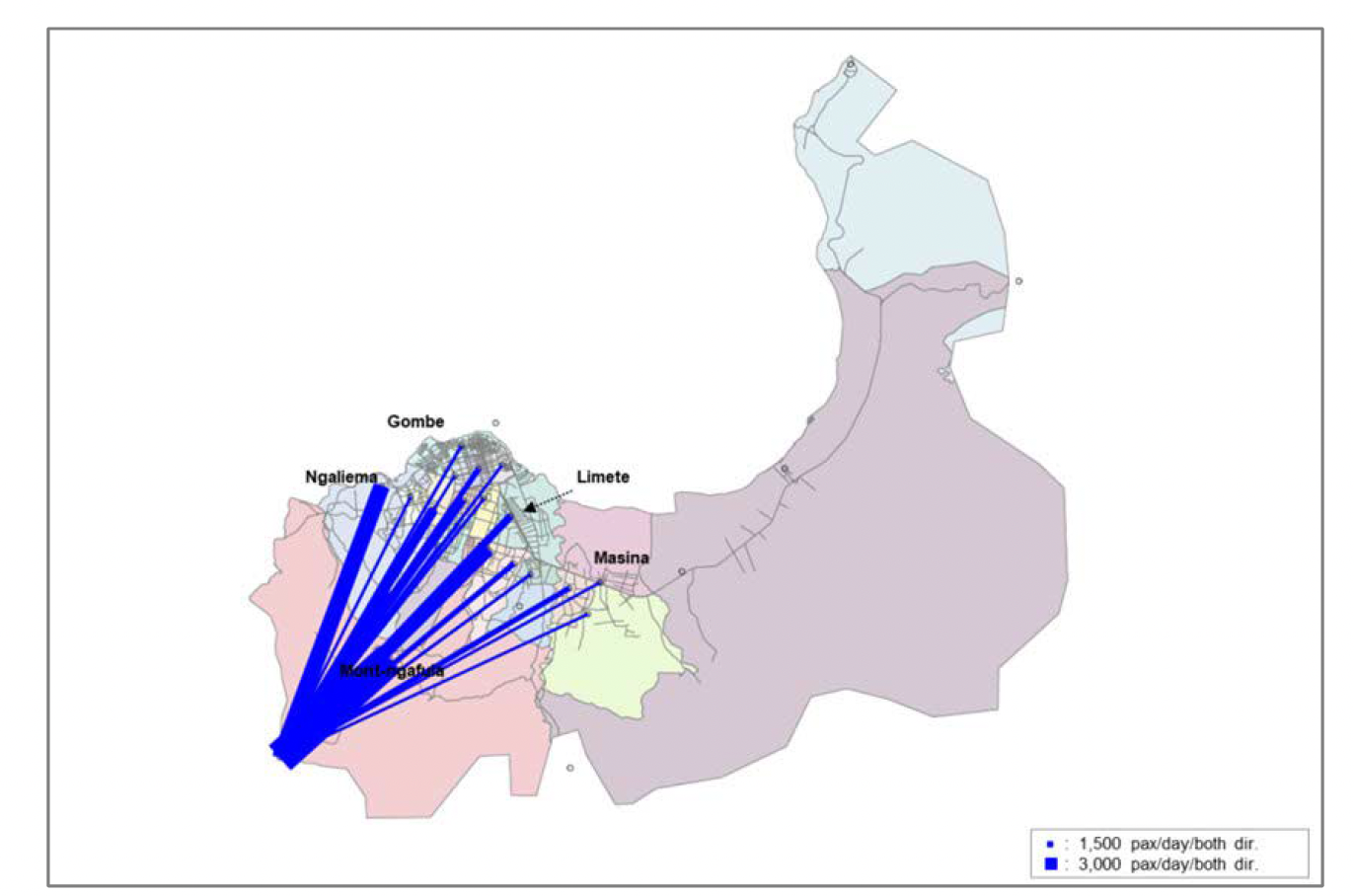}
        \caption{Traffic Main Origin Destination Pairs (Set 2)}
        \label{fig:traffic_data2}
    \end{subfigure}
    \hfill
    \begin{subfigure}[b]{0.32\linewidth}
        \centering
        \includegraphics[width=\linewidth]{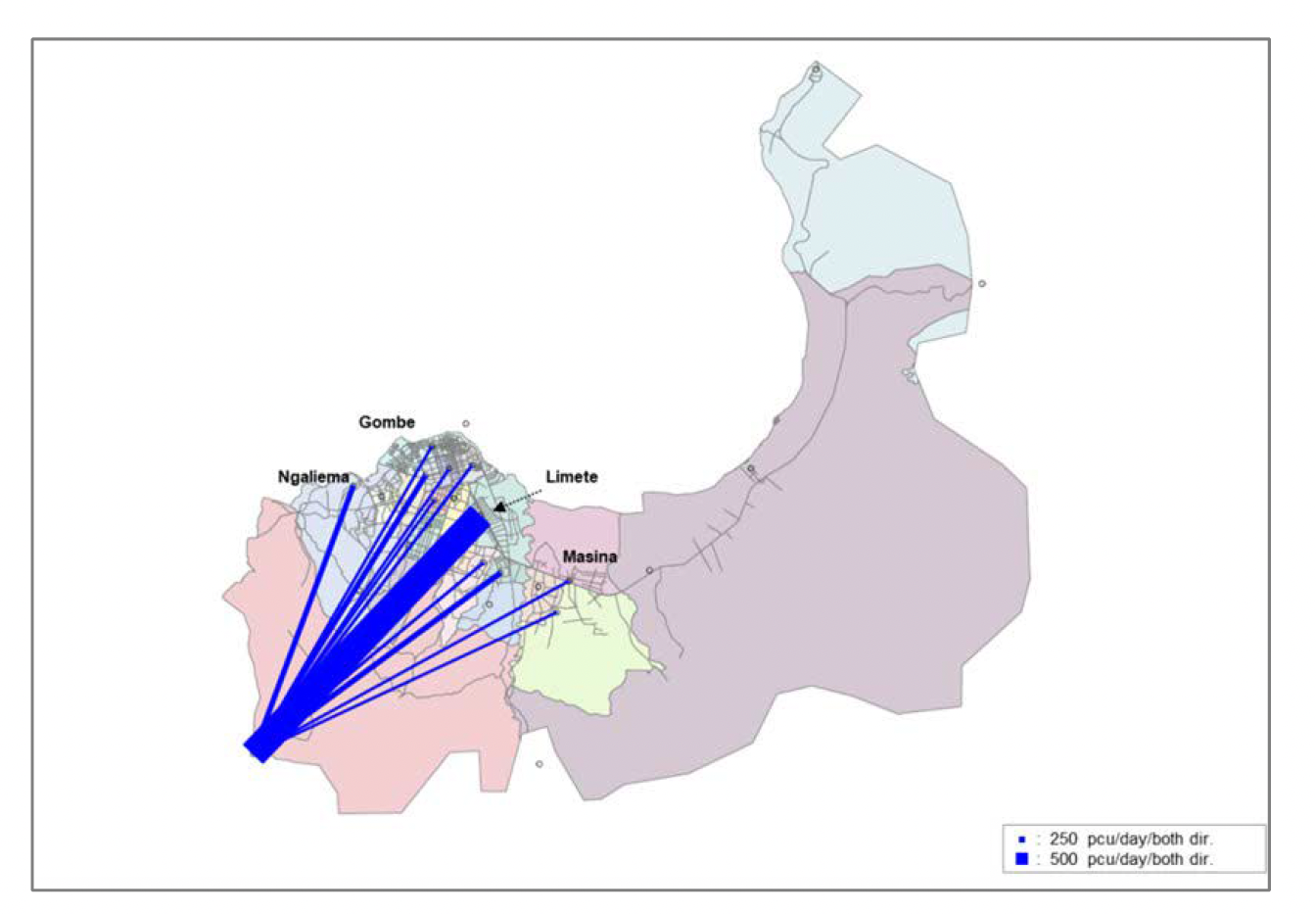}
        \caption{Traffic Main Origin Destination Pairs - Freight (Set 2b)}
        \label{fig:traffic_data3}
    \end{subfigure}
    \caption{Traffic volume data information in the Kinshasa road network (Freight and Passenger car units)}
    \label{fig:traffic_combined}
\end{figure}

\begin{figure}[H]
    \centering
    \begin{subfigure}[b]{0.32\linewidth}
        \centering
        \includegraphics[width=\linewidth]{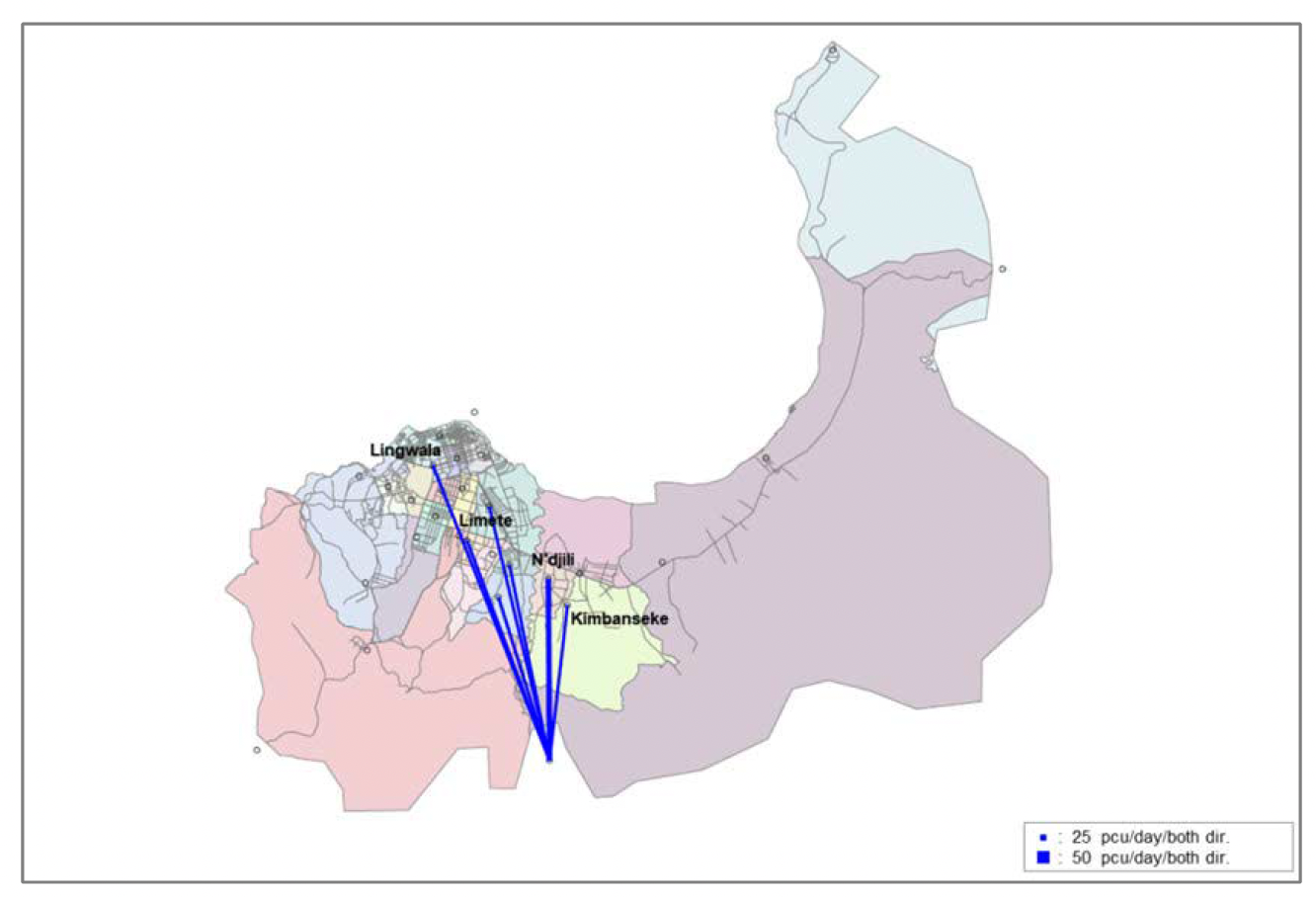}
        \caption{Traffic Main Origin Destination Pairs - Freight (Set 3)}
        \label{fig:traffic_data1b}
    \end{subfigure}
    \hfill
    \begin{subfigure}[b]{0.32\linewidth}
        \centering
        \includegraphics[width=\linewidth]{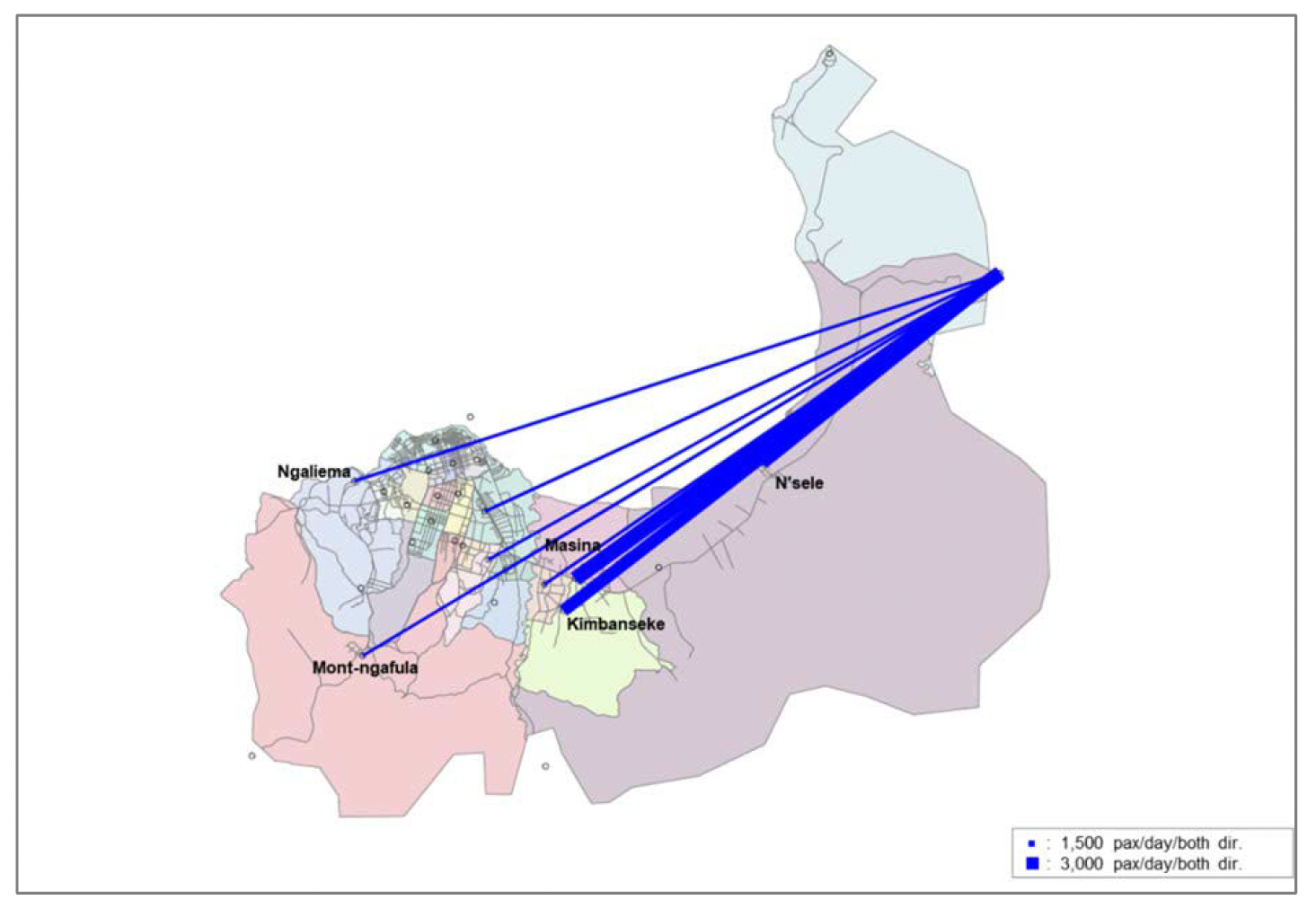}
        \caption{Traffic Main Origin Destination Pairs (Set 1)(Set 4)}
        \label{fig:traffic_data2b}
    \end{subfigure}
    \hfill
    \begin{subfigure}[b]{0.32\linewidth}
        \centering
        \includegraphics[width=\linewidth]{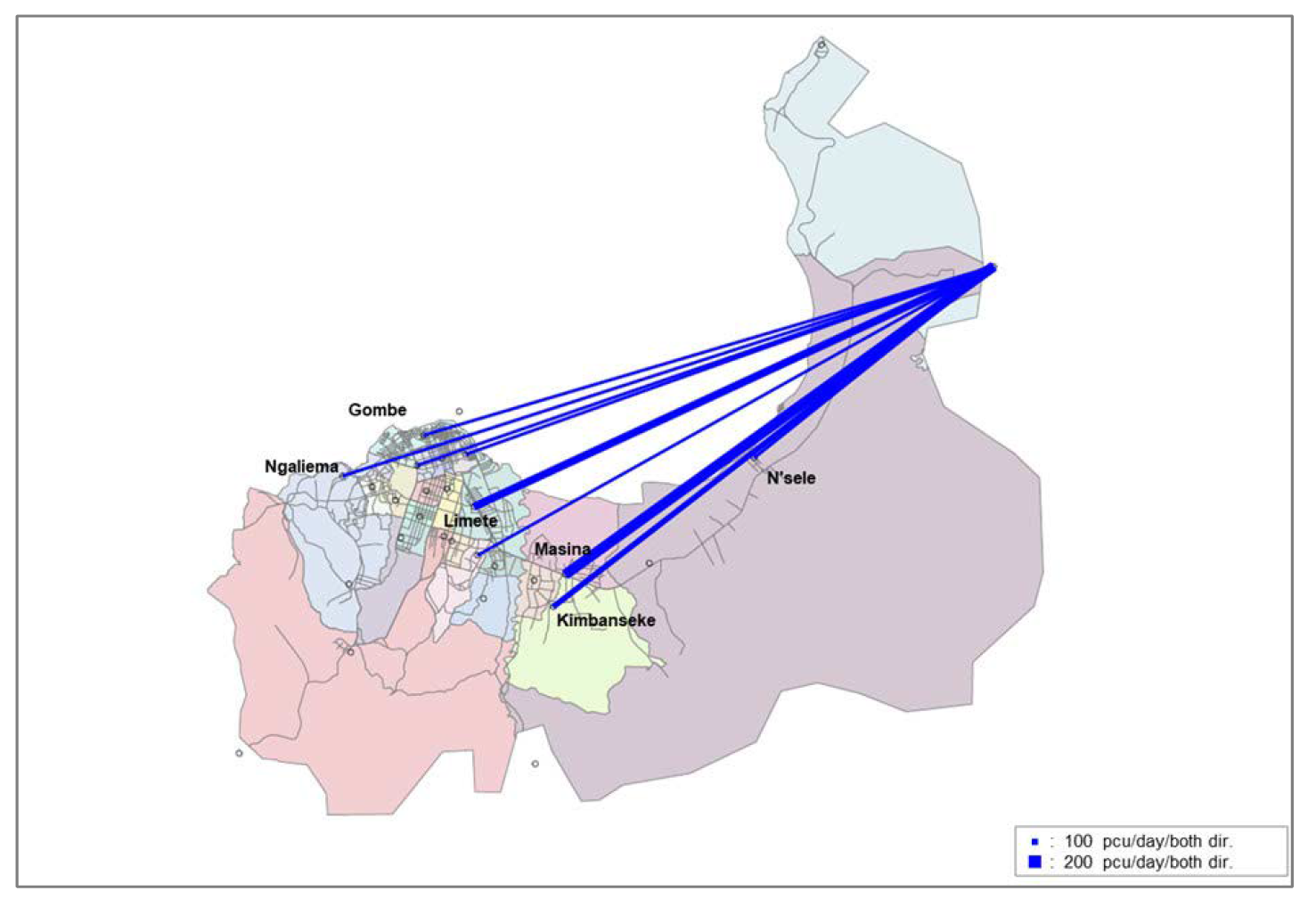}
        \caption{Traffic Main Origin Destination Pairs - Freight (Set 1) (Set 4b)}
        \label{fig:traffic_data3b}
    \end{subfigure}
    \caption{Traffic volume data information in the Kinshasa road network (Freight and Passenger car units)}
    \label{fig:traffic_combinedb}
\end{figure}

\begin{figure}[H]
    \centering
    \begin{subfigure}[b]{0.32\linewidth}
        \centering
        \includegraphics[width=\linewidth]{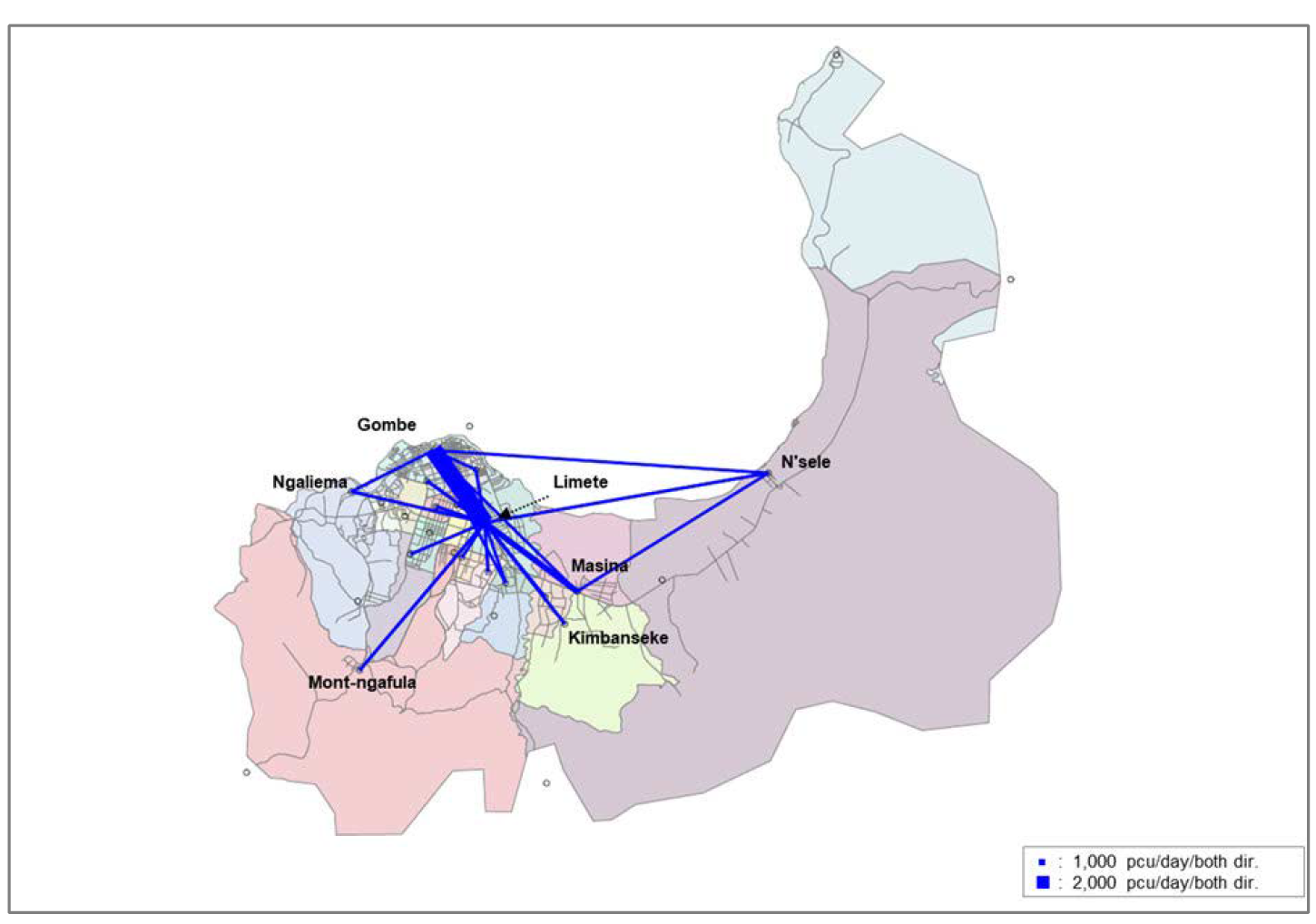}
        \caption{Traffic Main Origin Destination Pairs (Set 1) - Light good Trucks (Set 5a)}
        \label{fig:traffic_data1c}
    \end{subfigure}
    \hfill
    \begin{subfigure}[b]{0.32\linewidth}
        \centering
        \includegraphics[width=\linewidth]{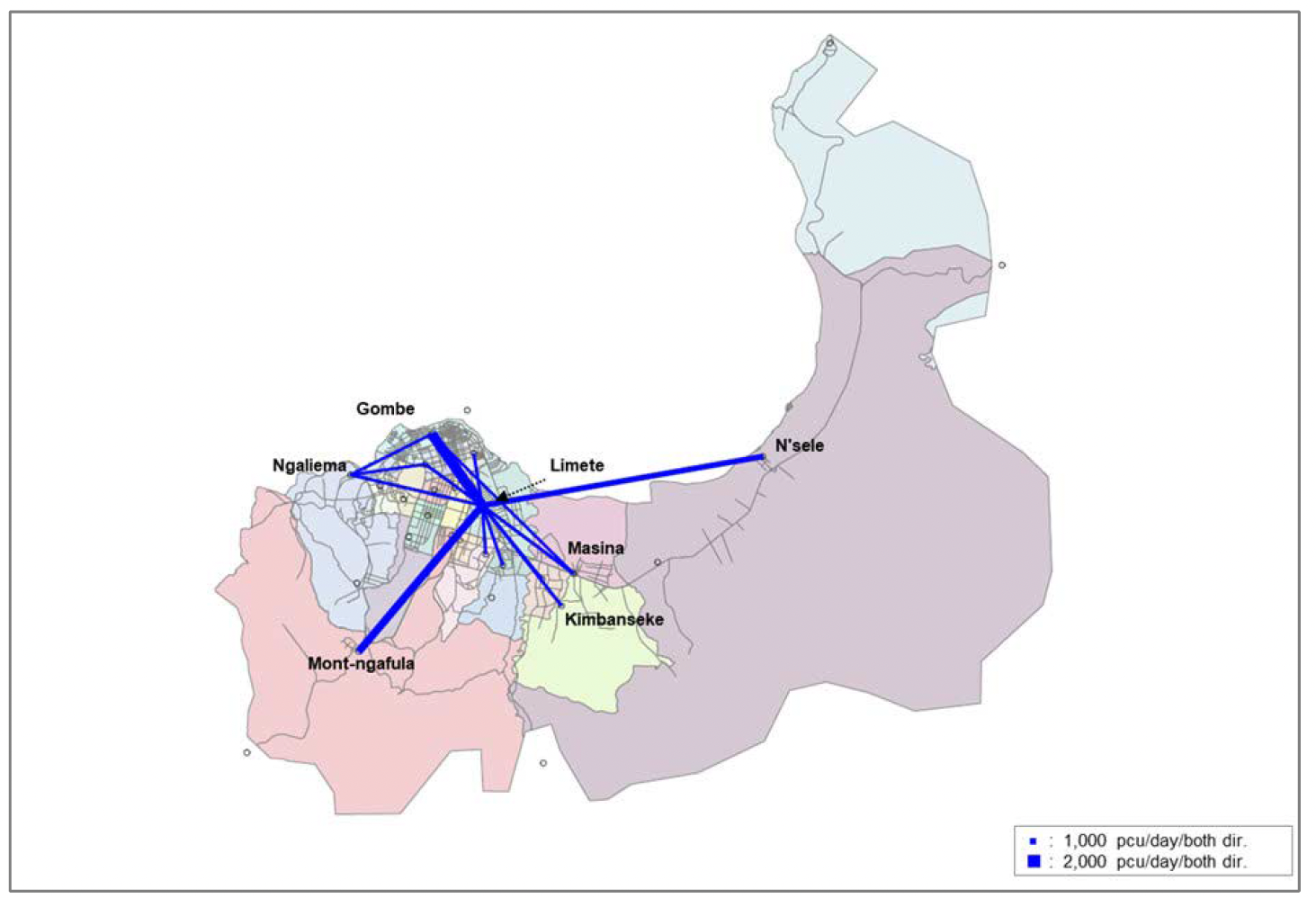}
        \caption{Traffic Main Origin Destination Pairs (Set 1) - Heavy Goods Trucks (Set 5b)}
        \label{fig:traffic_data2c}
    \end{subfigure}
    \hfill
    \begin{subfigure}[b]{0.32\linewidth}
        \centering
        \includegraphics[width=\linewidth]{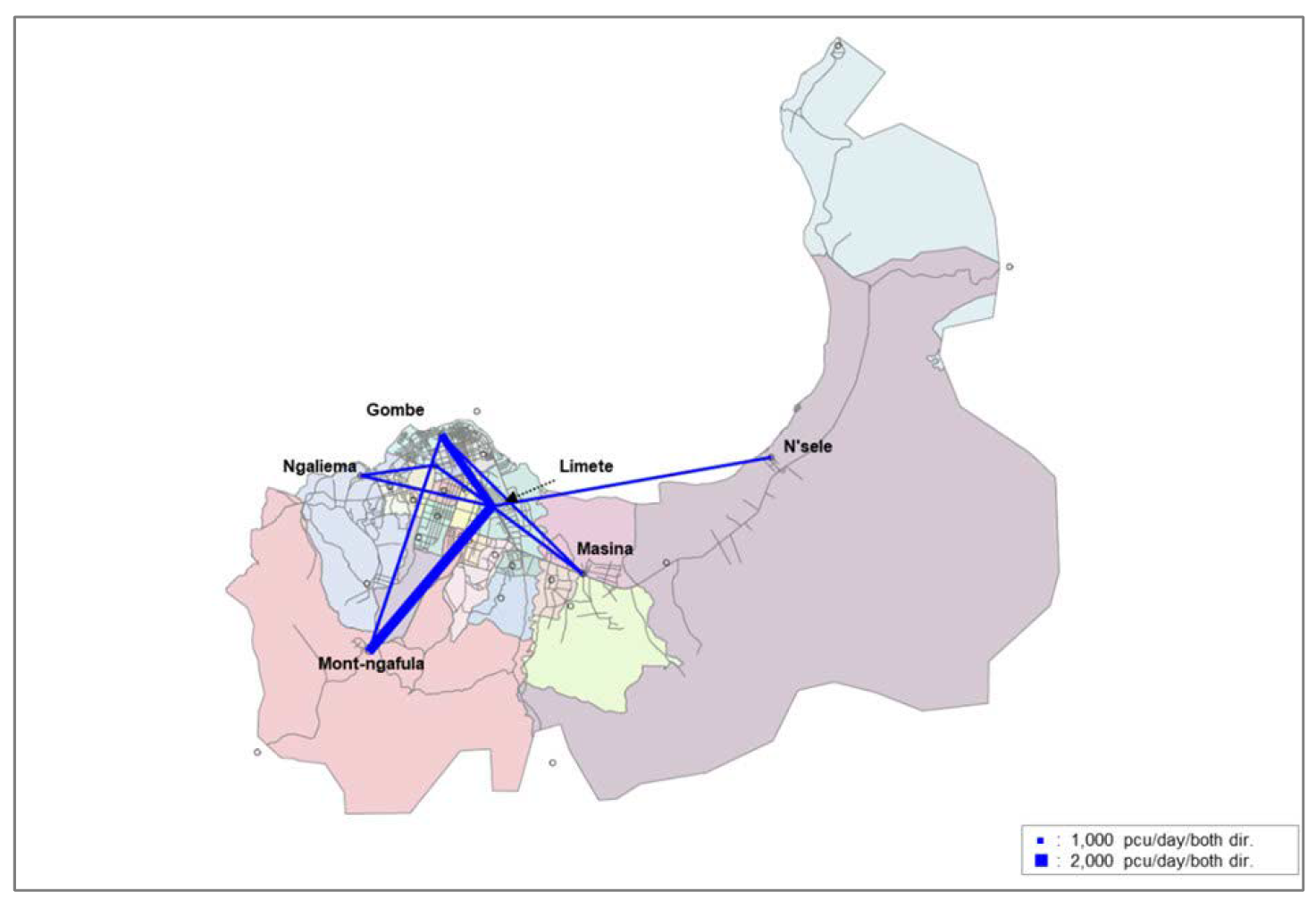}
        \caption{Traffic Main Origin Destination Pairs (Set 1) - Articulated Sets (Set 5c)}
        \label{fig:traffic_data3c}
    \end{subfigure}
    \caption{Traffic volume data information in the Kinshasa road network (Freight and Passenger car units)}
    \label{fig:traffic_combinedc}
\end{figure}

\begin{figure}[H]
    \centering
    \begin{subfigure}[b]{0.32\linewidth}
        \centering
        \includegraphics[width=\linewidth]{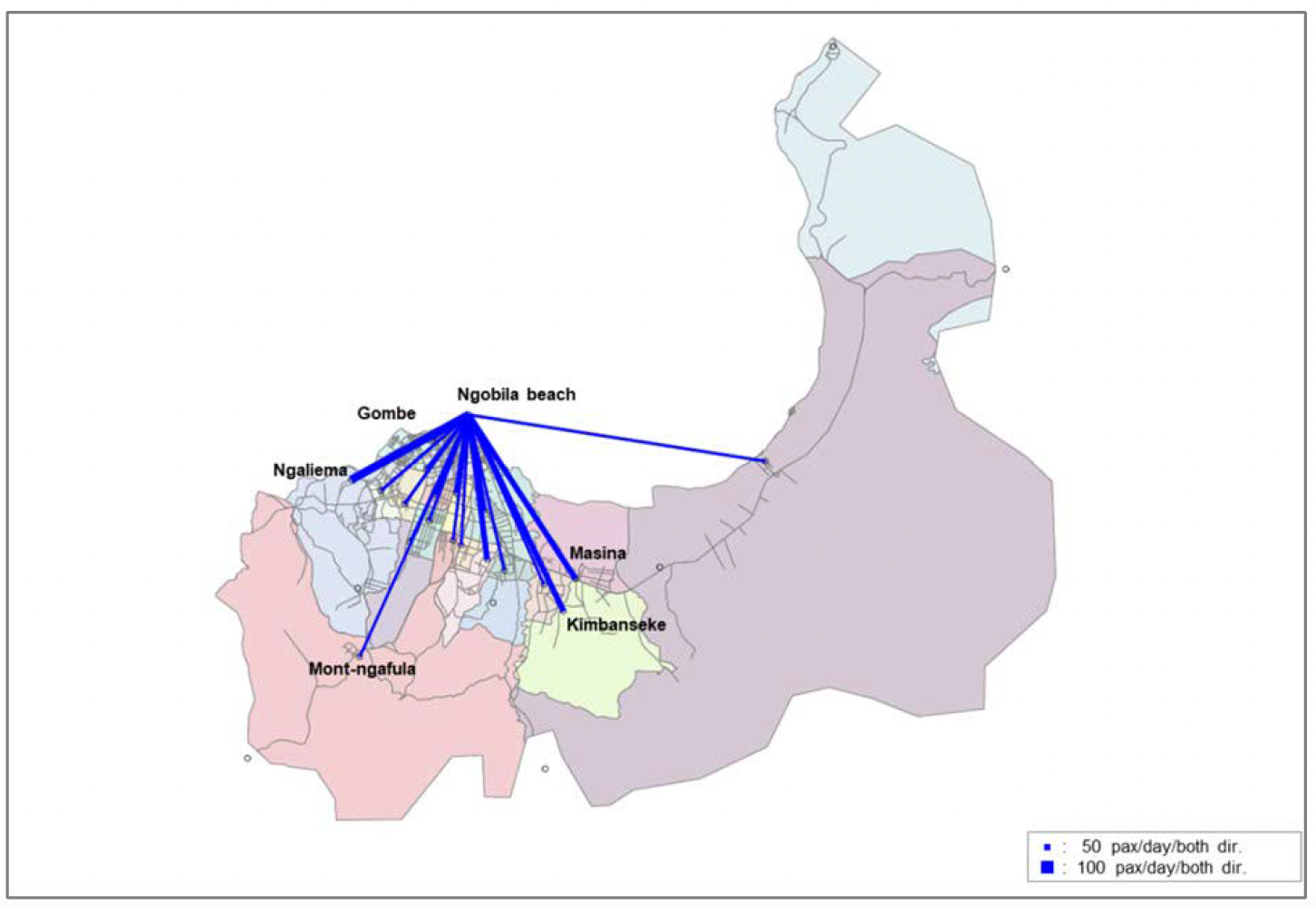}
        \caption{Traffic Main Origin Destination Pairs (Set 1) (Set 6)}
        \label{fig:traffic_data1d}
    \end{subfigure}
    \hfill

    \caption{Traffic volume data information in the Kinshasa road network (Freight and Passenger car units)}
    \label{fig:traffic_combinedd}
\end{figure}

\begin{figure}[h]
    \centering
    \includegraphics[width=0.7\linewidth]{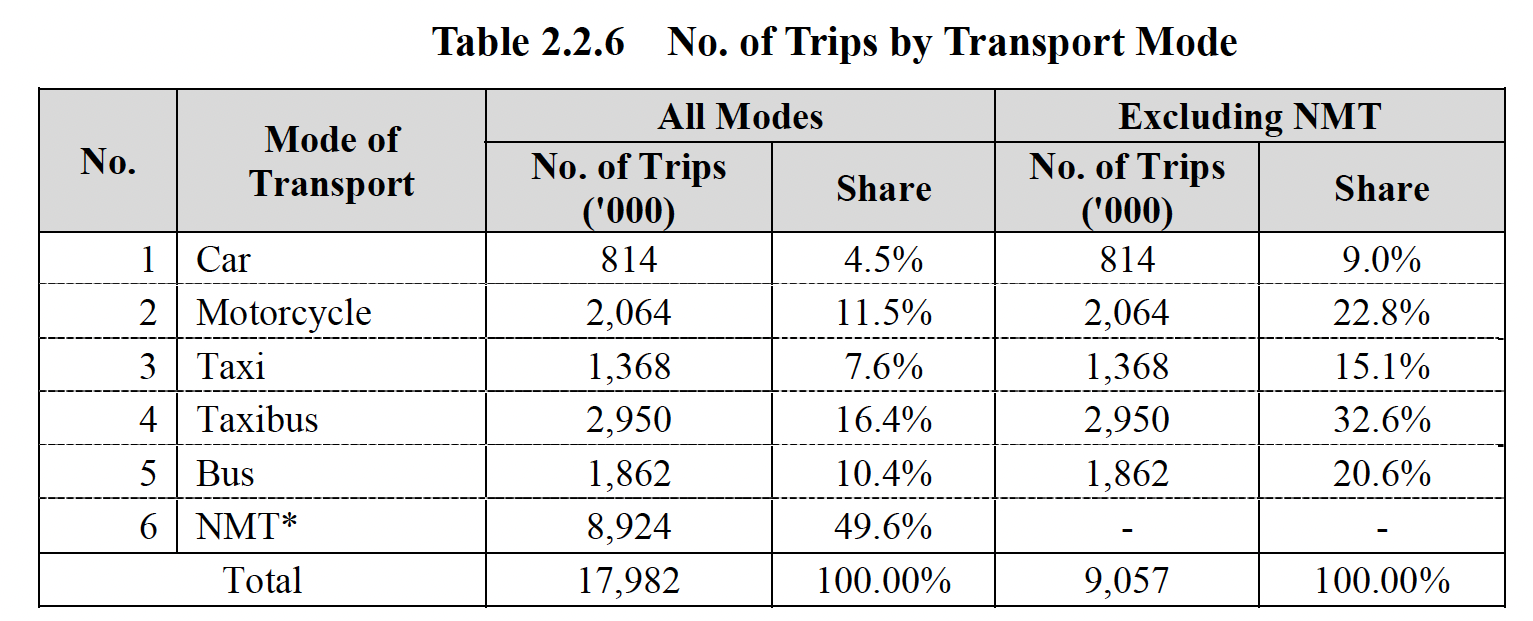}
    \caption{Number of trips per car unit}
    \label{fig:trip_car_unit}
\end{figure}

\noindent The number of passenger car units was converted using the formula 
\begin{equation}
    pcu = \sum_{i=1}^{ct}\alpha_i\frac{T}{\alpha_i'}
\end{equation}
\noindent where $T$ is the number of passengers, $\alpha_i$ is the number of occupants per transport type, and $\alpha_i$ is the trip share of each car type as given in Figure \ref{fig:trip_car_unit}.

\end{document}